\documentclass[final,3p,times,twocolumn]{elsarticle}

\usepackage{amssymb}

\usepackage{amsmath}
\usepackage{subcaption}
\usepackage{multirow}
\usepackage{pgfplots}
\pgfplotsset{compat=newest}
\usepackage{cite}
\usepackage{comment}
\newcommand{\conf}[1]{\scriptsize{$\pm$#1}}
\usepackage{xcolor}
\usepackage{soul}

\journal{Neural Networks}

\begin{document}

\begin{frontmatter}



\title{Hebbian Semi-Supervised Learning in a Sample Efficiency Setting\tnoteref{label1}}
\tnotetext[label1]{Published version: \texttt{doi.org/10.1016/j.neunet.2021.08.003}}


\author[inst1]{Gabriele Lagani\corref{cor1}}
\cortext[cor1]{Corresp. author. E-mail: \texttt{gabriele.lagani@phd.unipi.it}}
\affiliation[inst1]{organization={Computer Science Department, University of Pisa},
            city={Pisa},
            country={Italy}
            }

\author[inst2]{Fabrizio Falchi}
\author[inst2]{Claudio Gennaro}
\author[inst2]{Giuseppe Amato}

\affiliation[inst2]{organization={ISTI - CNR},
            city={Pisa},
            country={Italy}}

\begin{abstract}
We propose to address the issue of sample efficiency, in Deep Convolutional Neural Networks (DCNN), with a semi-supervised training strategy that combines Hebbian learning with gradient descent: all internal layers (both convolutional and fully connected) are pre-trained using an unsupervised approach based on Hebbian learning, and the last fully connected layer (the classification layer) is trained using Stochastic Gradient Descent (SGD). In fact, as Hebbian learning is an unsupervised learning method, its potential lies in the possibility of training the internal layers of a DCNN without labels. Only the final fully connected layer has to be trained with labeled examples.

We performed experiments on various object recognition datasets, in different regimes of sample efficiency, comparing our semi-supervised (Hebbian for internal layers + SGD for the final fully connected layer) approach with end-to-end supervised backprop training, and with semi-supervised learning based on Variational Auto-Encoder (VAE). The results show that, in regimes where the number of available labeled samples is low, our semi-supervised approach outperforms the other approaches in almost all the cases.


\end{abstract}


\end{frontmatter}



\section{Introduction}

In recent years, Deep Neural Networks (DNNs) have achieved impressive results in learning tasks such as computer vision \citep{he2016}, reinforcement learning \citep{silver2016}, natural language processing \citep{devlin2018}.


Today's neural networks are generally trained using Stochastic Gradient Descent (SGD) with the error backpropagation algorithm (backprop), which reaches high accuracy when a large number of labeled samples are available for training. However, gathering labeled samples is expensive, requires a significant amount of human work, and, in many applications, a large amount of training data is simply not available.

Therefore, researchers started to investigate strategies for sample efficient learning \citep{bengio2007, larochelle2009, weston2012, kingma2014, rasmus2015, zhang2016, chen2020}. In this setting, only a small number of labeled samples is assumed to be available. On the other hand, gathering unlabeled samples is relatively simple; therefore, these approaches exploit unlabeled samples to perform unsupervised training in addition to the supervised training process, leading to the so called \textit{semi-supervised} learning technique. It is well known that unsupervised pre-training helps initializing the network weights in the neighborhood of a good local optimum \citep{bengio2007, larochelle2009}, thus easing convergence in a successive supervised fine-tuning phase. Current semi-supervised approaches leverage autoencoder architectures for the unsupervised part of the task \citep{kingma2014, rasmus2015, zhang2016}, although they are still based on backprop. Another approach, SimCLR \citep{chen2020}, exploits data augmentation and an unsupervised contrastive criterion.

In this work, we address the sample efficiency issue by proposing 
%
%
a semi-supervised learning approach, where an initial unsupervised learning step, using all available data -- unlabeled and labeled (but without using label information) --, is followed by a supervised learning step using a small amount of labeled data. To perform the unsupervised learning step we explore the use of the Hebbian learning paradigm, which mimics more closely the synaptic adaptation mechanisms found in biological brains, according to neuroscientists. Hebbian learning is a local learning rule \citep{haykin, gerstner}, i.e. it does not require error backpropagation, and it does not require supervision. Moreover, the capabilities of biological brains to learn and generalize only from a limited number of labeled samples make this approach appealing for the sample efficient learning setting. Note also that backprop-based approaches are considered to be biologically implausible \citep{oreilly}.

The main contributions of this paper are the following:
\begin{itemize}
    \item We propose a semi-supervised learning approach that combines Hebbian learning with SGD on object recognition tasks with Deep Convolutional Neural Networks (DCNNs).
    \subitem - All available training samples, unlabeled and labeled, are used for an unsupervised Hebbian pre-training phase (without using label information), where a nonlinear Hebbian Principal Component Analysis (HPCA) learning rule is used to train internal layers (both convolutional and fully connected);
    \subitem - Then, labeled training samples and SGD are used to train a classifier, obtained as a final fully connected layer, on the features extracted from previous layers;
    \item The results are compared from a sample efficiency perspective with those obtained by a baseline network trained end-to-end with backprop, on the labeled samples, and with semi-supervised learning based on Variational Auto-Encoder (VAE) \citep{kingma2013} unsupervised pre-training, the latter using all the available samples (VAE-based semi-supervised learning was also the approach considered in \citep{kingma2014});
    \item Different datasets and different regimes of sample efficiency are explored, and it is shown that the proposed semi-supervised approach (Hebbian + SGD) outperforms the other approaches in almost all the cases where a limited number of labeled samples is available.
\end{itemize}

The remainder of this paper is structured as follows: 
Section~\ref{sec:rel_work} gives an overview on related work concerning semi-supervised training and Hebbian learning;
Section~\ref{sec:smpleff_scenario} illustrates the sample efficiency problem. Section \ref{sec:Hebbian-SGD} defines our approach to sample efficiency based on semi-supervised Hebbian + SGD learning;
Section~\ref{sec:background} provides a background on the Hebbian learning rule that we used in this work;
Section~\ref{sec:exp_setup} delves into the details of our experimental setup;
In Section~\ref{sec:results}, the results of our simulations are illustrated;
Finally, Section~\ref{sec:conclusions} presents our conclusions and outlines possible future developments.

\section{Related work} \label{sec:rel_work}

In this section, we present an overview of related work concerning both semi-supervised training and Hebbian learning.

\subsection{Semi-supervised training and sample efficiency}
Early work on deep learning had to face problems related to convergence to poor local minima during the training process, which led researchers to exploit a pre-training phase that allowed them to initialize network weights in a region near a good local optimum \citep{bengio2007, larochelle2009}. In these studies, greedy layerwise pre-training was performed by applying unsupervised autoencoder models layer by layer, thus training each layer to provide a compressed representation of the input for a successive decoding stage. It was shown that such pre-training was indeed helpful to obtain a good initialization for a successive supervised training stage.
In successive works, the idea of enhancing neural network training with an unsupervised learning objective was considered \citep{weston2012, kingma2014, rasmus2015, zhang2016}. In \citep{kingma2014}, Variational Auto-Encoders (VAE) are considered to perform an unsupervised pre-training phase using a limited amount of labeled samples. Also \citep{rasmus2015} and \citep{zhang2016} relied on autoencoding architectures to augment supervised training with unsupervised reconstruction objectives, showing that joint optimization of supervised and unsupervised losses helped to regularize the learning process. In \citep{weston2014}, joint supervised and unsupervised training was again considered, but the unsupervised learning part was based on manifold learning techniques. Another approach, SimCLR \citep{chen2020}, used a Contrastive Loss to perform the unsupervised learning part. The approach relies on data augmentation, in order to produce transformed variants of a given input, and the unsupervised loss basically imposes hidden representations to match for transformed variants generated from the same input.

\subsection{Hebbian learning}
Several variants of Hebbian learning rules were developed over the years, such as Hebbian learning with Winner-Takes-All (WTA) competition \citep{grossberg1976a}, Self-Organizing Maps \citep{kohonen1982}, Hebbian learning for Principal Component Analysis (PCA) \citep{haykin, sanger1989, karhunen1995, becker1996a}, Hebbian/anti-Hebbian learning \citep{pehlevan2015a, pehlevan2015c}. A brief overview is given in section \ref{sec:background}.
However, it was only recently that Hebbian learning started gaining attention in the context of DNN training \citep{wadhwa2016a, wadhwa2016b, bahroun2017, krotov2019, hwta, thesis}. In \citep{krotov2019}, a Hebbian learning rule based on inhibitory competition was used to train a neural network composed of fully connected layers on object recognition tasks. Instead, the Hebbian/anti-Hebbian learning rule developed in \citep{pehlevan2015a} was applied in \citep{bahroun2017} to extract convolutional features that were shown to be effective for classification. Convolutional layers were also considered in \citep{wadhwa2016a, wadhwa2016b}, where a Hebbian approach based on WTA competition was used to train the feature extractors. However, the previous approaches were based on relatively shallow network architectures (2-3 layers). A further step was taken in \citep{hwta, thesis}, where a Hebbian WTA learning rule was investigated for training a 6-layer Convolutional Neural Network (CNN). Also, a supervised variant of Hebbian learning was proposed to train the final classification layer. Hybrid network models were also considered, in which some layers where trained using backprop and others using Hebbian learning. The results suggested that Hebbian learning is suitable for training early feature detectors, as well as higher network layers, but not very effective for training intermediate network layers. Furthermore, Hebbian learning was successfully used to retrain the higher layers of a pre-trained network, achieving results comparable to backprop, but requiring fewer training epochs, thus suggesting potential applications in the context of transfer learning (see also \citep{magotra2019, magotra2020, canto2020}).

\section{Sample efficiency scenario} \label{sec:smpleff_scenario}

Training neural networks with supervision requires gathering a large amount of labeled training samples. This can be quite expensive, since it requires a considerable amount of human intervention for manually labeling collected samples. On the other hand, gathering unlabeled samples is generally considerably cheaper. In real world scenarios, one might need to solve AI problems for which only a small amount of labeled samples is available. In some cases, it is possible to start from a neural network pre-trained on a similar task, for which a good number of labeled samples was available, and just fine-tune it on the target task using the available labeled samples (thus performing transfer learning \citep{yosinski2014}. However, in many cases, the target task might be considerably different from the original task.


In such cases, it would be desirable to have a training algorithm that is capable of exploiting the latent information contained in large collections of unlabeled samples, while using only a small set of labeled samples for a successive stage of supervised training.


In fact, in applications where gathering large amounts of manually labeled data would be too expensive, it is often possible to collect a considerable amount of unlabeled samples at a relatively cheap cost. Therefore, it is desirable for an algorithm to be able to acquire knowledge about the data without using labels and learning to extract possibly useful features by, for example, learning to distinguish frequent shapes and patterns, or detecting rarely occurring anomalies. However, a fully supervised, backprop-based, approach, typically requires many labeled samples to achieve a good performance.


Formally, the sample efficiency learning scenario can be stated as follows: let $\mathcal{T}$ be our \textit{training set}, let $s \in \mathcal{T}$ be an element (called \textit{training sample}) of the training set. 
Elements of $\mathcal{T}$ are drawn from a statistical distribution with \textit{probability density function (pdf)} $p(s)$:
\begin{equation}
    s \; \sim \; p(s)
\end{equation}
Let $N = |\mathcal{T}|$ be the number of training samples in $\mathcal{T}$ (where $|\cdot|$ denotes the cardinality of the set). 
Let $\mathcal{L}$ be another set, whose elements $l \in \mathcal{L}$ are called \textit{labels}. 
Let 
\begin{equation}
    \phi : \mathcal{T} \to \mathcal{L}
\end{equation}
be a function that maps training samples to a corresponding label. 
This function is unknown, except for a subset of $\mathcal{T}$ for which labels are given. Specifically, we define the \textit{labeled set} as a subset $\mathcal{T}_L \subset \mathcal{T}$ of elements for which the image of the function $\phi$ is known: 
\begin{equation}
    \mathcal{T}_L = \{s \in \mathcal{T} \; | \; \phi (s) \; \text{is known}\}
\end{equation}
Let's define the \textit{unlabeled set} $\mathcal{T}_U$ as the complement of $\mathcal{T}_L$ w.r.t. $\mathcal{T}$. 
Therefore, for the unlabeled set, label information is not available; nonetheless, samples from $\mathcal{T}_U$ are drawn from the same statistical distribution as the samples from $\mathcal{T}_L$ and $\mathcal{T}$, i.e. from $p(s)$. 
In a \textit{sample efficiency} scenario, the number of samples in $\mathcal{T}_L$ is typically much smaller than the total number of samples $N$ in $\mathcal{T}$, i.e. 
\begin{equation}
    |\mathcal{T}_L| << |\mathcal{T}|
\end{equation}
In particular, an $r \, \%$-\textit{sample efficiency} scenario is characterized by 
\begin{equation}
    |\mathcal{T}_L| = \frac{r}{100}|\mathcal{T}|
\end{equation}
i.e. the size of the labeled set is $r \, \%$ that of the whole training set (labeled plus unlabeled). 
A neural network is required to approximate, by a training process, the function $\phi$. 
For a given $\mathcal{T}$ and a given $r \, \%$-sample efficiency regime, we define a neural network to \textit{perform better} than another if it reaches higher accuracy in mapping samples to correct labels (according to function $\phi$), given that both networks are trained using $\mathcal{T}_L$ and $\mathcal{T}_U$ splits that are compliant with the considered $r \, \%$-sample efficiency scenario, i.e. both networks have been trained with a number of labeled samples equal to $r \, \%$ of the total number of samples in $\mathcal{T}$. 
The training process can be \textit{supervised}, in which case only samples from $\mathcal{T}_L$, and the associated labels, are used, or it can be \textit{unsupervised}, in which case all samples from both $\mathcal{T}_U$ and $\mathcal{T}_L$ are used, but without using the labels for the latter.

\section{Hebbian-SGD Semi-Supervised learning approach}
\label{sec:Hebbian-SGD}

Traditional supervised approaches based on backpropagation work well provided that the size of the labeled set is sufficiently large, but they do not exploit the unlabeled set. 
In the semi-supervised approach that we propose, we aim to learn an approximation of function $\phi$ by means of a two stage process: first, we run an unsupervised training stage in which we use samples from both $\mathcal{T}_U$ and $\mathcal{T}_L$ (but without using the labels for the latter), in order to learn latent representations for samples drawn from $p(s)$; subsequently, we use the small number of available labeled samples from $\mathcal{T}_L$ (with the corresponding labels) in a supervised training phase. 
During the first phase, latent representations are obtained from hidden layer of a DCNN, which are trained using an unsupervised, biologically plausible, Hebbian learning algorithm. 
During the second phase, supervised training is applied on a final linear classifier, by running a SGD optimization procedure using only the few labeled samples at our disposal (with the corresponding labels).

\begin{figure}[t]
\centering
\includegraphics[width=0.3\textwidth]{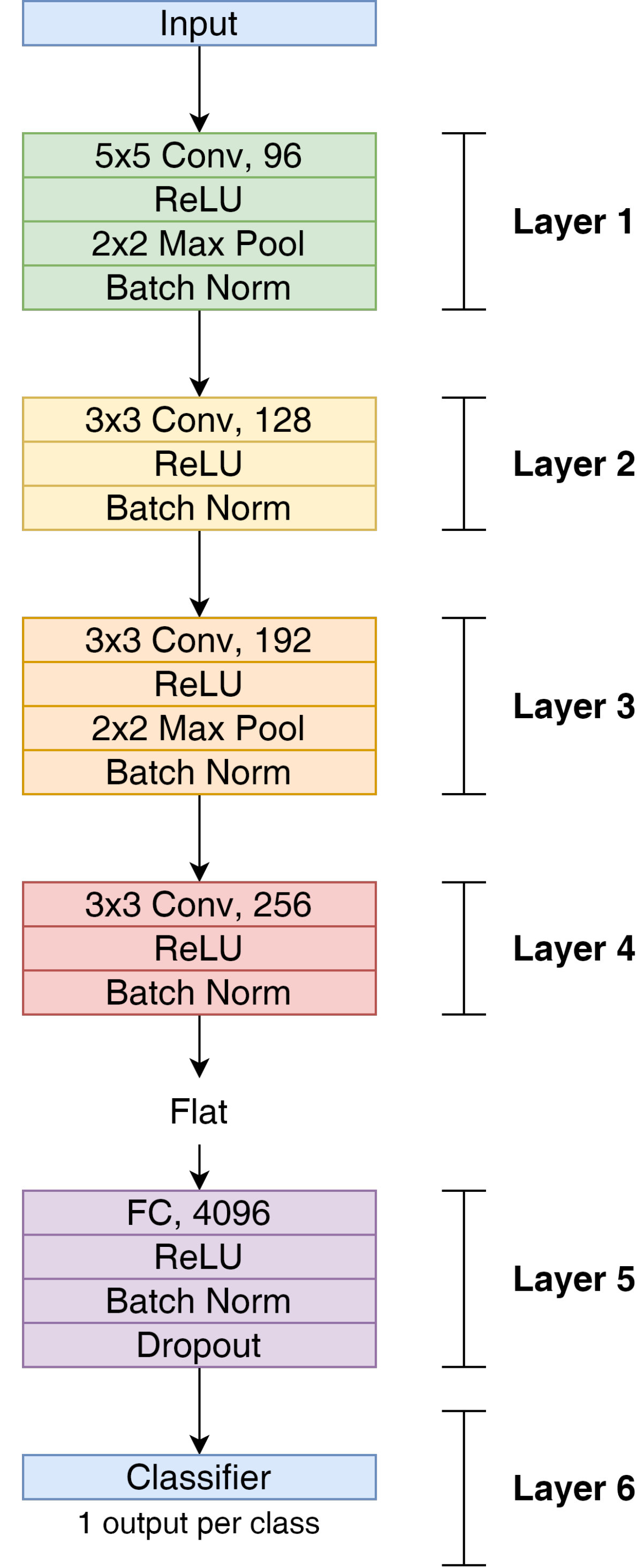}
\caption{The neural network used for the experiments.}
\label{fig:network}
\end{figure}

Specifically, we propose the following semi-supervised learning approach, which combines Hebbian learning with SGD, applied to a deep convolutional neural network architecture as the one in Figure \ref{fig:network}:

\begin{enumerate}
    \item the internal layers of the neural network are trained without supervision using the Hebbian learning rule, given in Eq. \ref{eq:lrn_rule} discussed in Section \ref{sec:background}, and all the available samples from both $\mathcal{T}_U$ and $\mathcal{T}_L$ (but without using the labels for the latter), independently from the final fully connected layer, used as classifier;
    \item the final fully connected layer, used as classifier, is trained with SGD, freezing the weights of previous layers, and using only the few labeled examples in $\mathcal{T}_L$ (with the corresponding labels);
    \item We also considered the case in which, during the supervised phase, the weights of the previous layers are not frozen, but they are adapted together with the final classifier, thus performing end-to-end fine tuning on the Hebbian pre-trained network; we compare the results of the Hebbian approach with and without fine tuning, in order to be able to tell apart the contributions of Hebbian learning and end-to-end fine tuning on the final result.
\end{enumerate}

As we will show in the following, we found that our semi-supervised approach achieves interesting results in low sample efficiency regimes. In fact, when only a small number of labeled samples is used (roughly from 1\% to 5\% of the total number of training samples contained in the considered datasets), our approach generally offers better results than full backprop training, and VAE-based semi-supervised learning. Thus it represents a promising solution to real life applications, where the number of available labeled samples is small.


In the experiments discussed below, we compared the performance of our approach with full backprop, and with VAE-based semi-supervised learning, considering various levels of sample efficiency. Moreover, we run different experiments in which the final classifier was placed on top of a different inner layer of the network, in order to evaluate the quality of the feature representations extracted at different network depths in the classification task.
As we will show later, our approach performs better than the other methods in almost all the cases when we consider low sample efficiency scenarios, where the percentage of labeled samples is between 1\% and 5\% (i.e. $r \, \%$-sample efficiency scenarios with $1 \leq r \leq 5$). Further improvements come when end-to-end fine tuning is used, in addition to Hebbian pre-training.

\section{Background on Hebbian learning} \label{sec:background}

For a given neuron, the weight updates according to the Hebbian learning rule, in its most basic form, can be expressed as \citep{haykin}:
\begin{equation}
    \mathbf{w}_{new} = \mathbf{w}_{old} + \Delta \mathbf{w}
\end{equation}
where $\mathbf{w}_{new}$ is the updated weight vector, $\mathbf{w}_{old}$ is the old weight vector, and $\Delta \mathbf{w}$ is the weight update. The latter term is computed as follows:
\begin{equation}
    \Delta \mathbf{w} = \eta \, y \, \mathbf{x}
\end{equation}
Here, $\mathbf{x}$ is the input vector of the neuron, $\mathbf{w}$ is its weight vector, $y = \mathbf{w^T x}$ is its output, and $\eta$ is the learning rate. According to this rule, the weight on a given synapse is increased when the input on that synapse and the output of the neuron are simultaneously high, so that correlation between simultaneously active neurons is reinforced.

\begin{figure*}
    \centering
    \subfloat[Update step]{
        \includegraphics[width=0.4\textwidth]{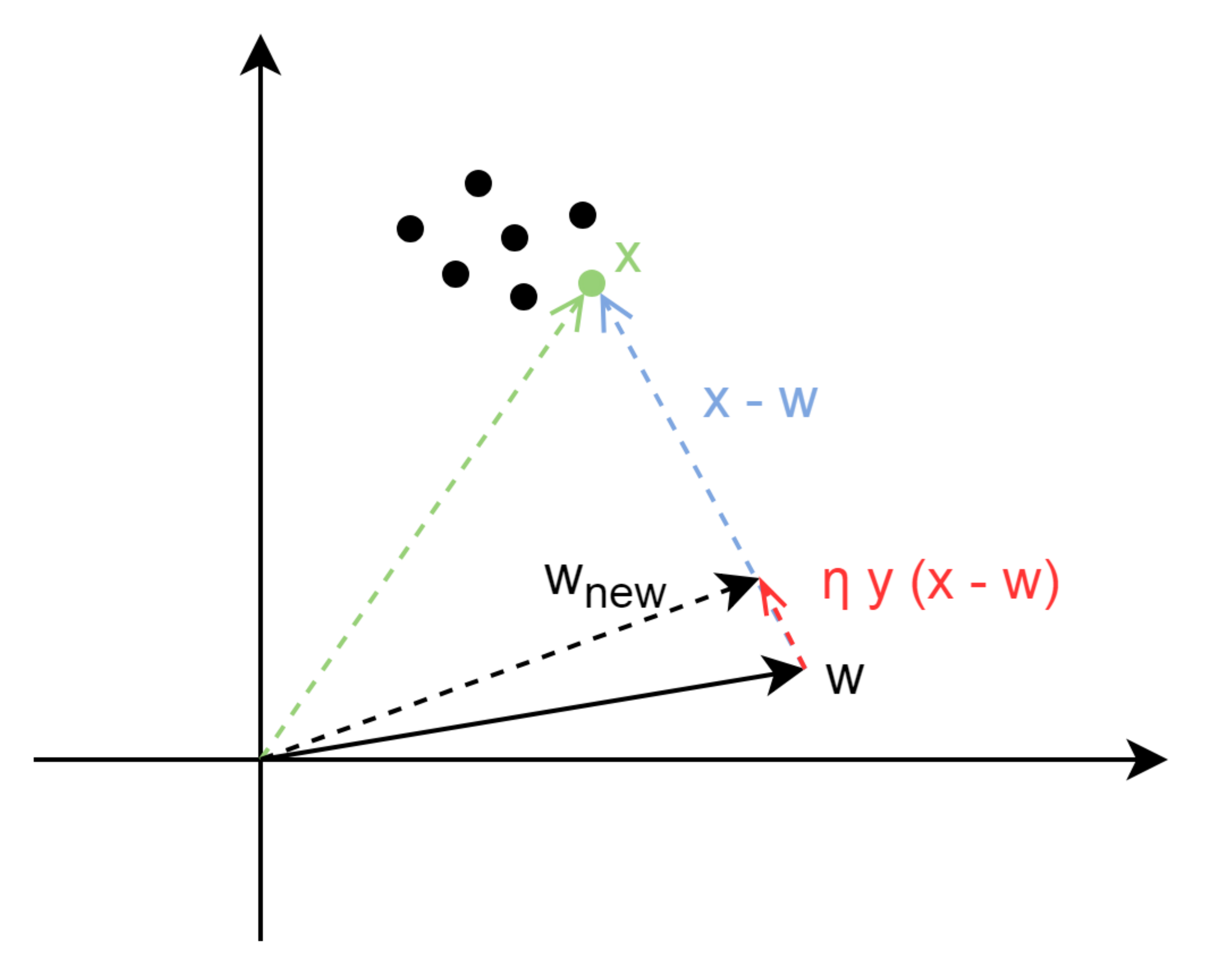}
    }
    ~
    \subfloat[Final position after convergence]{
        \includegraphics[width=0.4\textwidth]{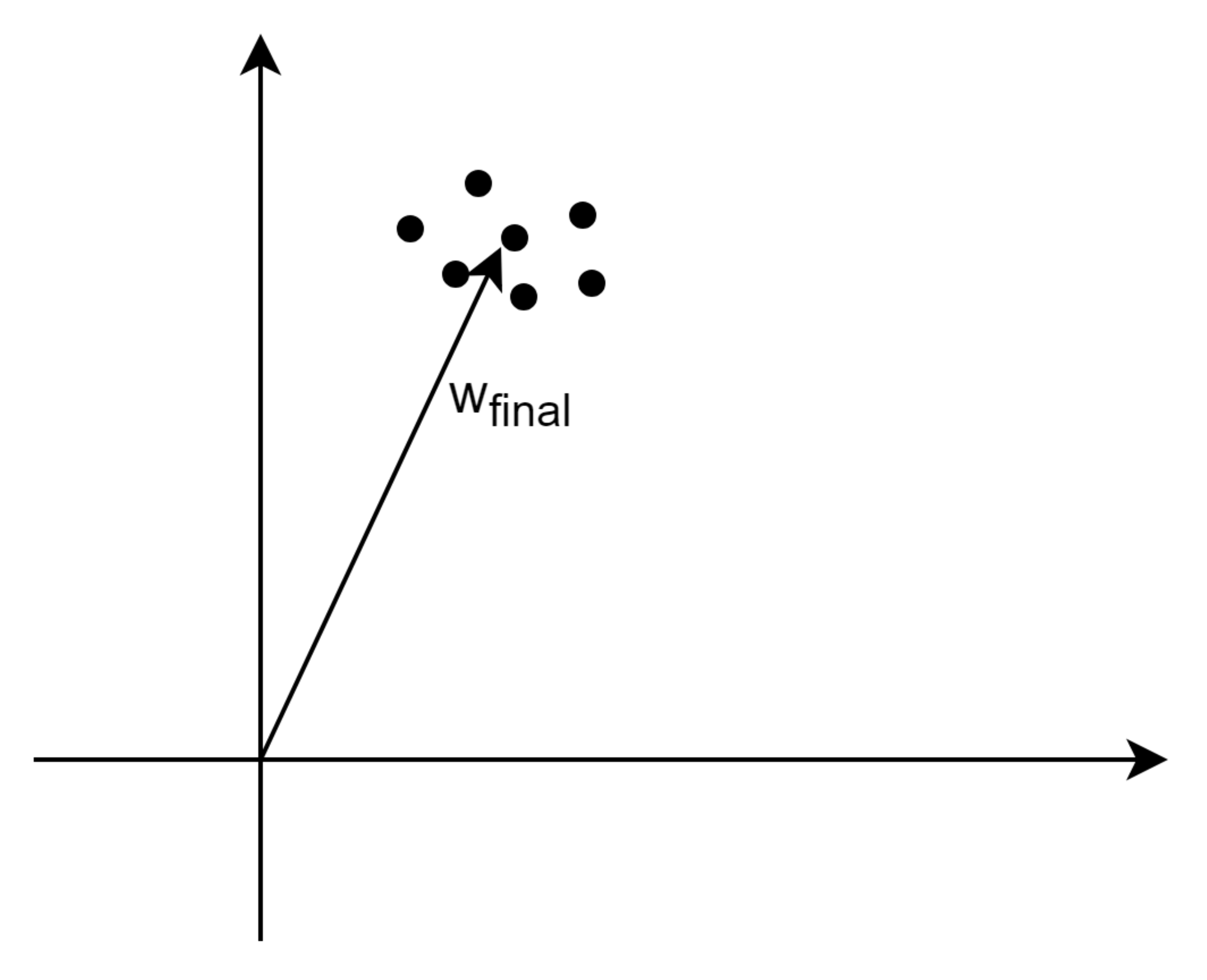}
    }
    \caption{Hebbian updates with weight decay.}
    \label{fig:hebb_update_wd}
\end{figure*}

The learning rule above is unstable, in that neuron weights are allowed to grow unbounded. To prevent this circumstance, a weight decay term is typically added: $\Delta \mathbf{w} = \eta \, y \, \mathbf{x} - \lambda \, \mathbf{w}$. In particular, the term $\lambda$ can be chosen in the form $\lambda = \eta \, y$, as suggested, for example, in works on competitive learning \citep{grossberg1976a, kohonen1982}, leading to the following expression for $\Delta \textbf{w}$:
\begin{equation} \label{eq:hebb_wta}
    \Delta \mathbf{w} = \eta \, y \, \mathbf{x} - \eta \, y \, \mathbf{w} = \eta \, y \, (\mathbf{x} - \mathbf{w})
\end{equation}
The rule above can be easily interpreted intuitively: when an input vector is presented to the neuron, its vector of weights takes a small step towards the input, so that the neuron will exhibit a stronger response if a similar input is presented again in the future. When a series of inputs drawn from a cluster are presented to the neuron, the weight vector converges towards the cluster center (Figure \ref{fig:hebb_update_wd}).

\begin{figure*}
    \centering
    \subfloat[Update step]{
        \includegraphics[width=0.5\textwidth]{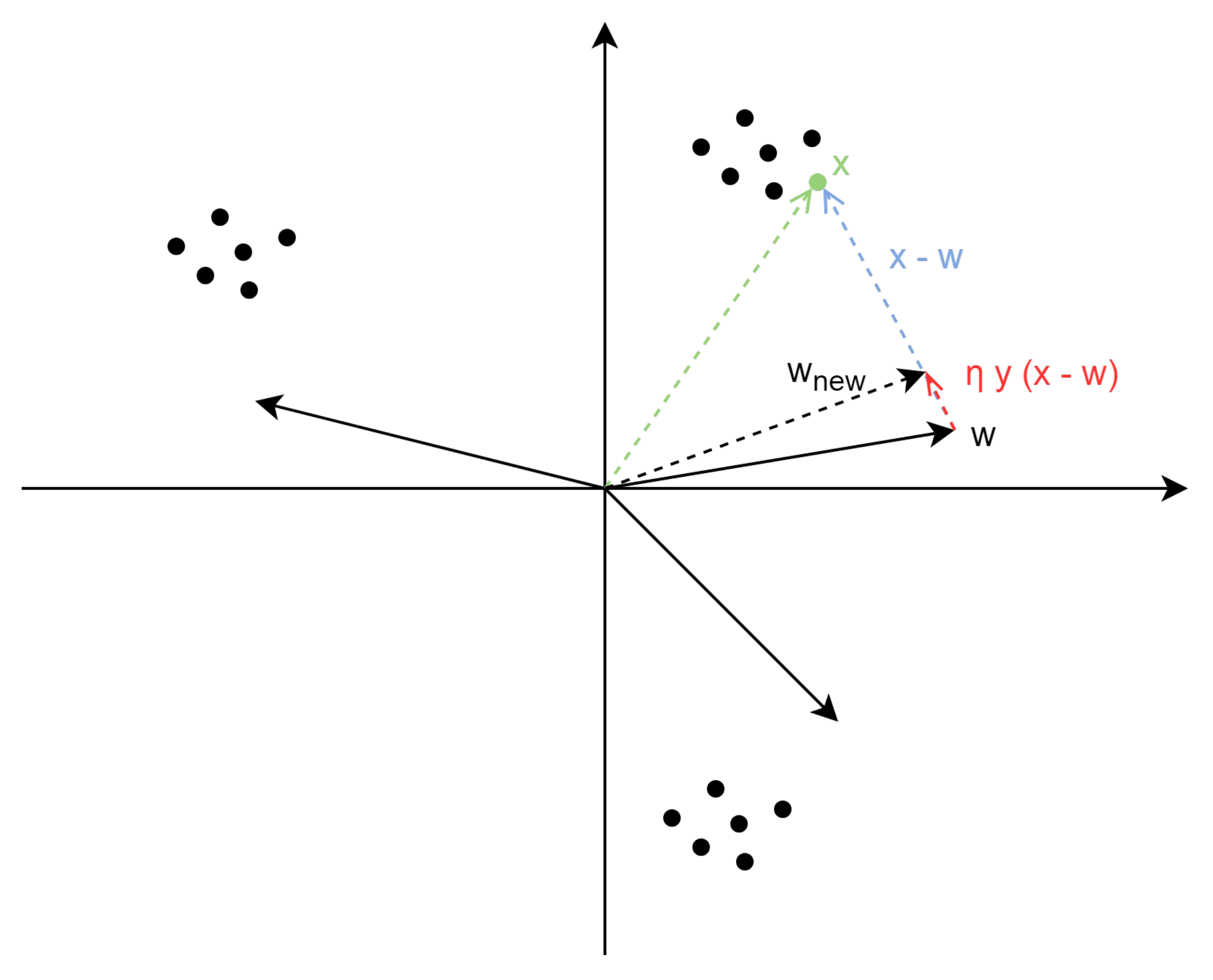}
    }
    ~
    \subfloat[Final position after convergence]{
        \includegraphics[width=0.5\textwidth]{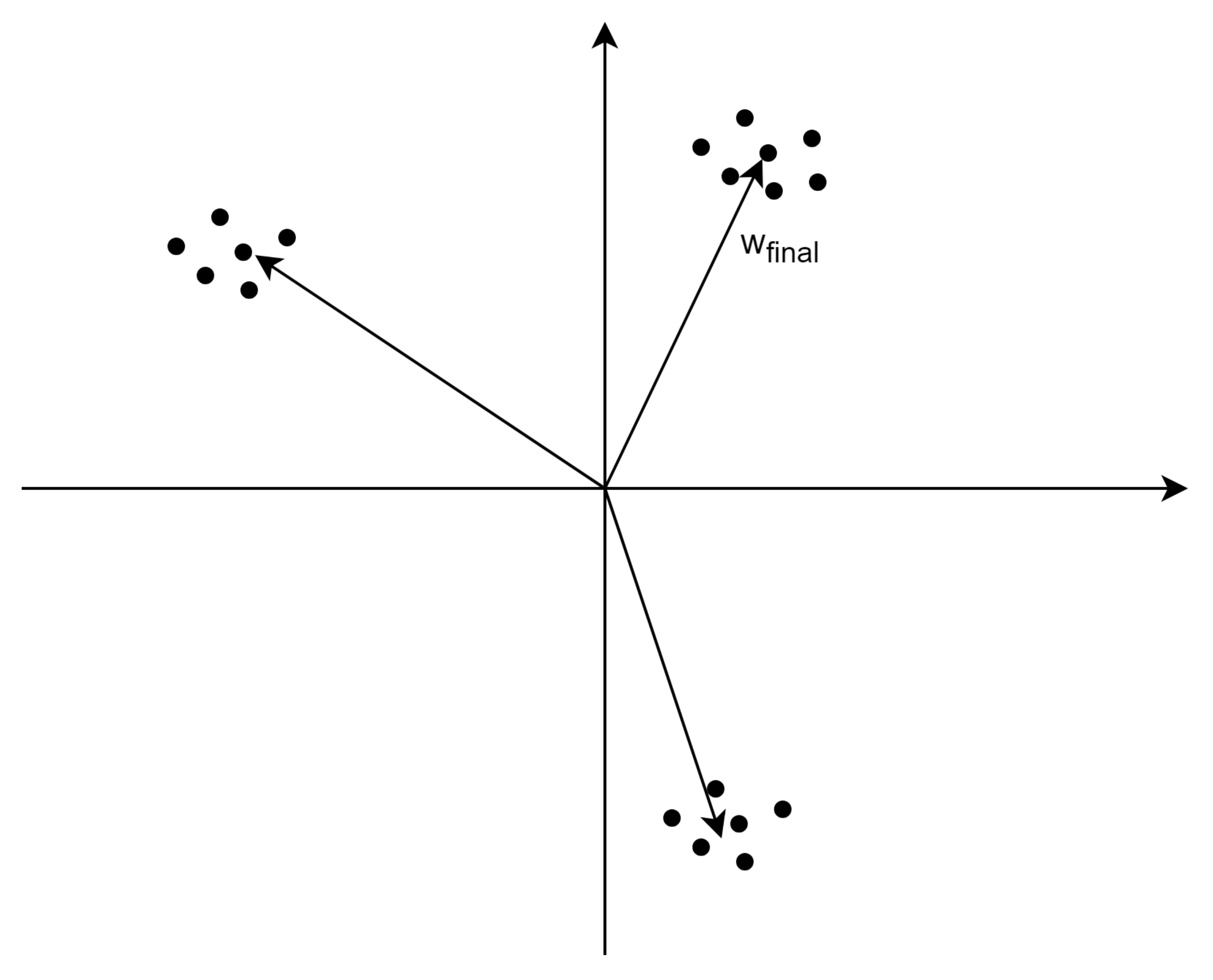}
    }
    \caption{Hebbian updates with Winner-Takes All competition.}
    \label{fig:hebb_update_wta}
\end{figure*}

In a complex neural network, several neurons are involved in representing the input. Ideally, we would like each neuron to learn to represent a different property of the inputs. Therefore, we need a strategy to prevent neurons from learning redundant information. The Winner-Takes-All (WTA) \citep{grossberg1976a} strategy was proposed for this purpose. It was motivated by the observation that biological neural networks exhibit inhibitory competition, i.e. when a neuron fires, it provokes the inhibition of neighboring neurons. In the WTA strategy, when an input is presented to a neural network layer, a \textit{winner} neuron is determined. This is the one whose weight vector is closest to the current current input, and it is the only neuron allowed to perform a weight update according to Eq. \ref{eq:hebb_wta}. In this way, if a similar input is presented again in the future, the same neuron will be more likely to win again. At the same time, different neurons are induced to represent different properties of the inputs, namely the centroids of the clusters formed by the input data points (Fig.~\ref{fig:hebb_update_wta}).

The WTA provides a \textit{quantized} encoding scheme, in which one particular neuron activates to encode a particular input. However, it has been argued that a distributed type of encoding would be more powerful \citep{foldiak1990, olshausen1996a}, in which different neurons simultaneously activate to encode different properties of the input.

A more distributed coding scheme can be achieved if neuron weight vectors are trained to encode the principal components of the input data. This can be achieved by Hebbian Principal Component Analysis (HPCA) learning rules, in an efficient, online manner \citep{haykin, sanger1989, becker1996a, karhunen1995}. The Hebbian PCA learning rule is obtained by minimizing the \textit{representation error} loss function, defined as:
\begin{equation} \label{eq:repr_err}
    L(\mathbf{w_i})  = E[(\mathbf{x} - \sum_{j=1}^i y_j \, \mathbf{w_j})^2]
\end{equation}
where, in this case, the subscript $i$ is used to denote the $i^{th}$ neuron in the layer and $E[\cdot]$ is the mean value operator.
Let's give an intuitive explanation to the latter equation. Each weight vector represents a pattern that is stored by a neuron. The neuron's output represents ``how much" of that pattern was found in the current input. The summation in the equation represents a \textit{reconstruction} of the current input, obtained as a linear combination of the weights of the neurons, with coefficients given by the corresponding neuron outputs. Finally, we compute the squared error between the actual input $\mathbf{x}$ and the reconstruction, which represents how much the current reconstruction is far from the actual input. Therefore, minimizing this loss corresponds to finding network weights that produce the most accurate reconstruction. 
The objective in Eq. \ref{eq:repr_err} reduces to classical PCA, in the case of linear neurons and zero centered data, which requires to maximize output variance, while the weight vectors are subject to orthonormality constraints, as shown in \citep{haykin, sanger1989, becker1996a, karhunen1995}. 
In the following, we assume that the input data are centered around zero. If this is not true, we just need to subtract the average $E[\mathbf{x}]$ from the inputs beforehand.

Minimizing the objective in Eq. \ref{eq:repr_err} leads to the following Hebbian PCA learning rule \citep{haykin, sanger1989, becker1996a, karhunen1995}:
\begin{equation} 
    \Delta \mathbf{w_i} = \eta y_i (\mathbf{x} - \sum_{j=1}^i y_j \mathbf{w_j})
\end{equation}
In particular, we consider the case in which neurons have nonlinear activation functions $f()$, so that the representation error becomes:
\begin{equation}
    L(\mathbf{w_i})  = E[(\mathbf{x} - \sum_{j=1}^i f(y_j) \, \mathbf{w_j})^2]
\end{equation}
Minimization of this objective \citep{karhunen1995, becker1996a} leads to the nonlinear Hebbian PCA learning rule that we used in our experiments:
\begin{equation} \label{eq:lrn_rule}
    \Delta \mathbf{w_i} = \eta f(y_i) (\mathbf{x} - \sum_{j=1}^i f(y_j) \mathbf{w_j})
\end{equation}

\section{Experimental setup} \label{sec:exp_setup}

In the following, we describe the details of our experiments and comparisons, discussing the network architecture and the training procedure\footnote{The code to reproduce the experiments described in this paper is available at: \\ \texttt{https://github.com/GabrieleLagani/HebbianPCA/tree/hebbpca}.}.

\subsection{Datasets used for the experiments}

The experiments were performed on the following datasets: CIFAR10, CIFAR100 \citep{cifar} and Tiny ImageNet \citep{tinyimagenet}. 

The CIFAR10 dataset contains 50,000 training images and 10,000 test images, belonging to 10 classes. Moreover, the training images were randomly split into a training set of 40,000 images and a validation set of 10,000 images.

The CIFAR100 dataset also contains 50,000 training images and 10,000 test images, belonging to 100 classes. Also in this case, the training images were randomly split into a training set of 40,000 images and a validation set of 10,000 images.

The Tiny ImageNet dataset contains 100,000 training images and 10,000 test images, belonging to 200 classes. Moreover, the training images were randomly split into a training set of 90,000 images and a validation set of 10,000 images.

We considered a range of sample efficiency regimes, i.e. we assumed that only a limited number of labeled samples was available for training. We considered the following sample-efficiency regimes: the amount of labeled samples were respectively 1\%, 2\%, 3\%, 4\%, 5\%, 10\%, 25\% and 100\% of the whole training set. This corresponds to 400, 800, 1,200, 1,600, 2,000, 4,000, 10,000, 40,000 labeled samples  for the CIFAR10 and CIFAR100 datasets, and to 900, 1,800, 2,700, 3,600, 4,500, 9,000, 22,500, 90,000 labeled samples for Tiny ImageNet.

\subsection{Network architecture and training}

We considered a six layer neural network as shown in Fig. \ref{fig:network}: five deep layers plus a final linear classifier, obtained as a fully connected layer on top of previous layers. The various layers were interleaved with other processing stages (such as ReLU nonlinearities, max-pooling, etc.). The architecture was inspired by AlexNet \citep{krizhevsky2012}, but with slight modifications in order to reduce the overall computational cost of training.

For each sample efficiency regime, we trained a deep network with our semi-supervised approach, using the Hebbian PCA (HPCA) rule in Eq. \ref{eq:lrn_rule} (in which the nonlinearity was set to the ReLU function) in the internal layers, followed by the SGD training step in the classification layer only (plain HPCA), or also fine tuning previous network weights (HPCA plus Fine Tuning -- HPCA+FT).

For each sample efficiency configuration we also created a baseline for comparison, training an identical network using SGD with error backpropagation in all layers, in a supervised, end-to-end fashion. We considered both a network trained from scratch, and a network pre-trained with unsupervised VAE method \citep{kingma2013} and then fine-tuned by supervised backprop. VAE-based semi-supervised learning was also the method considered in \citep{kingma2014}.

As already stated, while supervised training only uses the labeled samples, the unsupervised training step (for the internal layers) uses the entire dataset, consisting of both labeled and unlabeled samples (although the label information is not used). 

\subsection{Testing sample efficiency at different layer depths}

\begin{figure}[t]
\centering
\includegraphics[width=\linewidth]{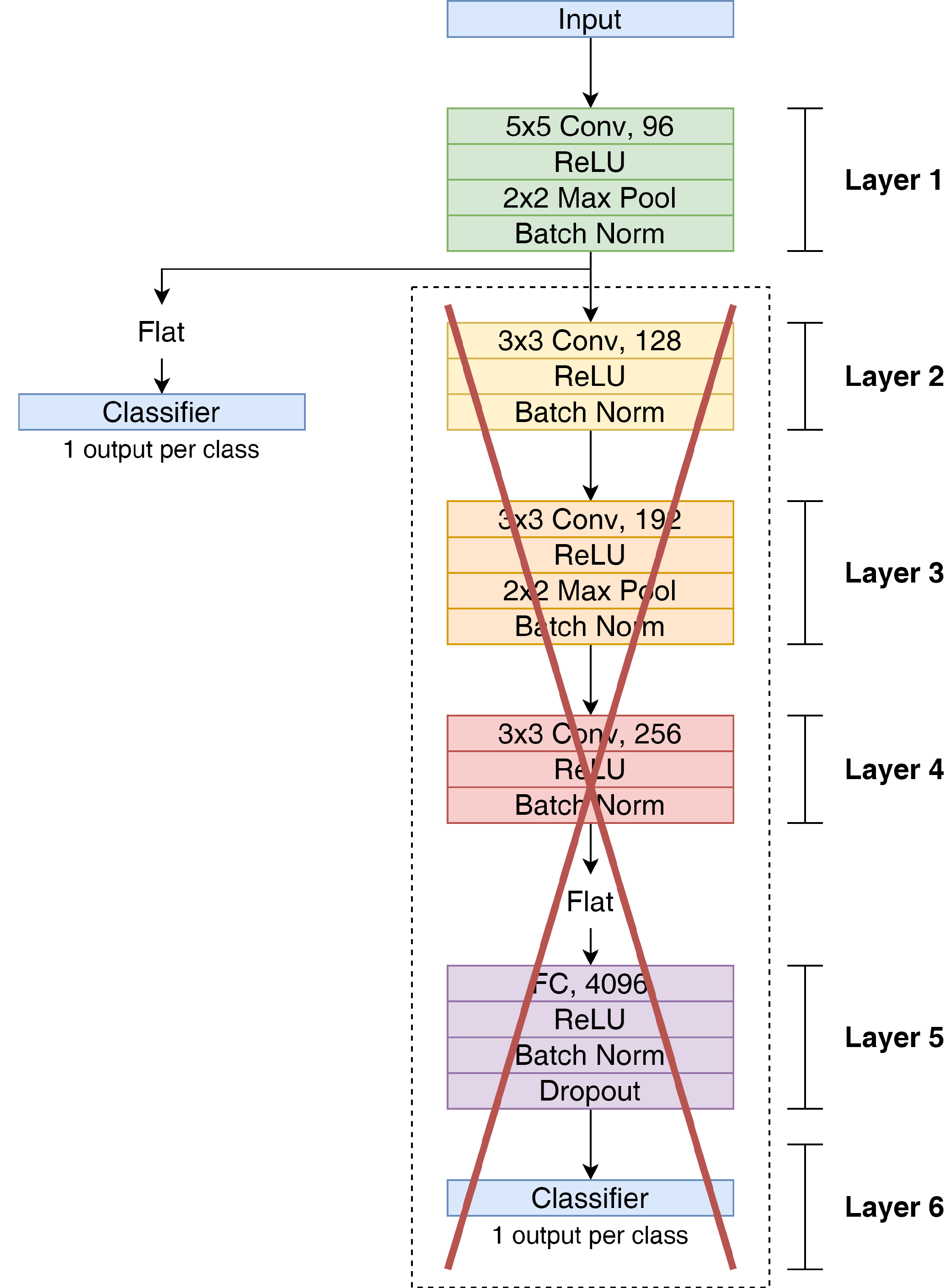}
\caption{A neural network is cut in correspondence of layer 1, and a linear classifier is placed on top of the features extracted from that layer, in order to evaluate their quality in classification tasks.}
\label{fig:network_classifier_on_conv1}
\end{figure}

In our experiments, in addition to evaluate the entire network trained as discussed above, we also evaluated the sample efficiency capability on the various internal layers of the trained models. To this end, we cut the networks in correspondence of the output of the various layers and we trained a new linear classifier on top of each already pre-trained layer (for instance, Fig. \ref{fig:network_classifier_on_conv1} shows a classifier placed on top of the features extracted from the first layer), and the resulting accuracy was evaluated. These new classifiers were trained with supervision using SGD, hence using only the limited number of labeled training samples given by the sample efficiency regime considered.
This process was done for the HPCA trained network with previous layers frozen, for the HPCA trained network with fine tuning of previous layers (HPCA+FT), for the SGD network trained from scratch with supervised backprop, and for the SGD network pre-trained by VAE and then fine tuned with supervised backprop, in order to make comparisons. More details are given below.

\subsection{Details of training}

We implemented our experiments using PyTorch. Training was performed in 20 epochs using mini-batches of size 64. Networks were fed input images of size 32x32 pixels. Experiments were performed using five different seeds for the Random Number Generator (RNG), averaging the results and computing 95\% confidence intervals.

\begin{figure*}
\centering
\includegraphics[width=0.8\textwidth]{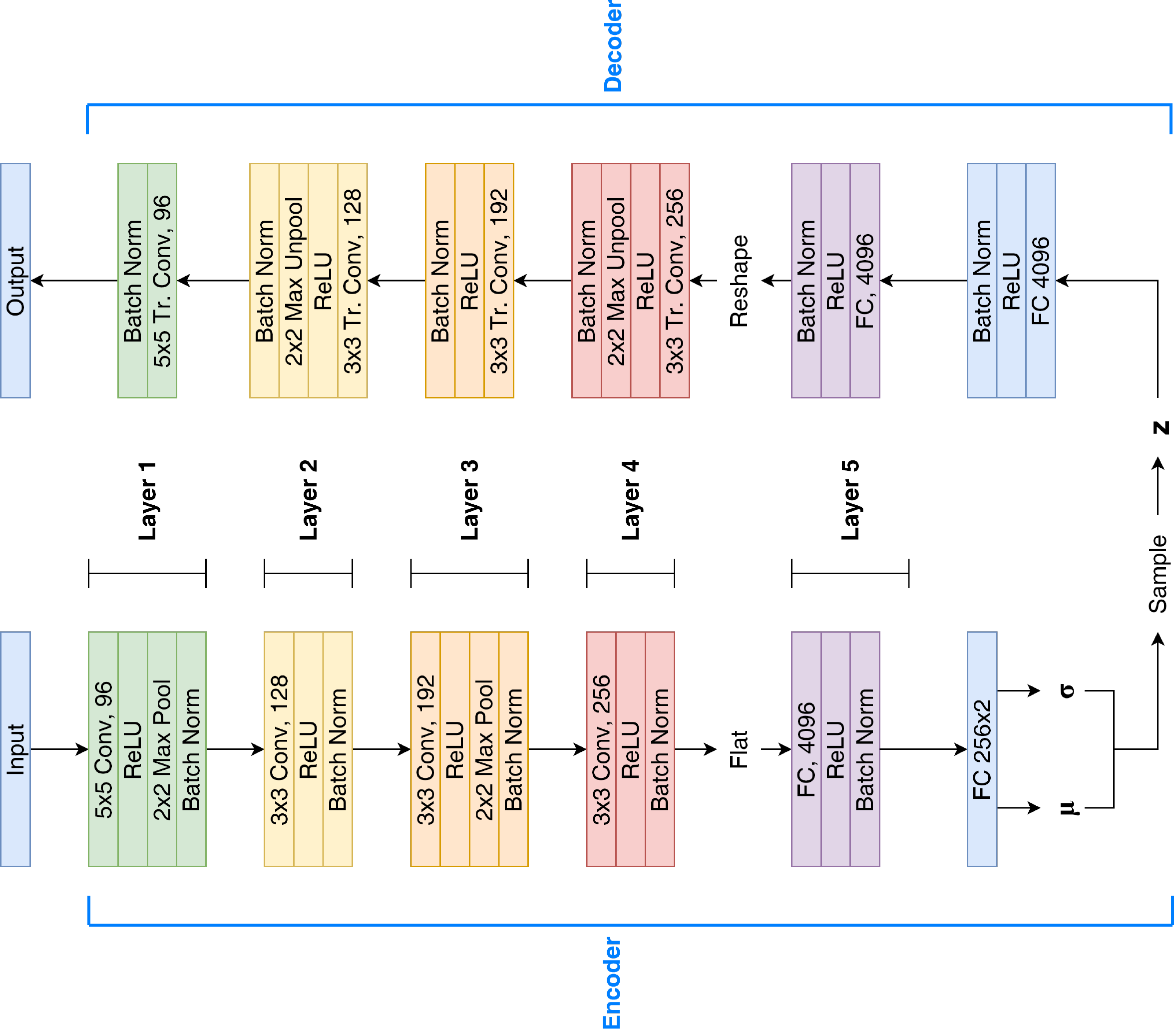}
\caption{VAE architecture used in our experiments.}
\label{fig:vae}
\end{figure*}

For SGD training of the baseline network, the initial learning rate was set to $10^{-3}$ and kept constant for the first ten epochs, while it was halved every two epochs for the remaining ten epochs. We also used momentum coefficient $0.9$, Nesterov correction, dropout rate 0.5 and L2 weight decay penalty coefficient set to $5 \cdot  10^{-2}$ for CIFAR10, $10^{-2}$ for CIFAR100 and $5 \cdot 10^{-3}$ for Tiny ImageNet. Cross-entropy loss was used as optimization metric. When VAE pre-training was used, the network in Fig. \ref{fig:network}, up to layer 5, acted as encoder, with an extra layer mapping layer 5 output to 256 gaussian latent variables, while a specular network branch acted as decoder (Fig. \ref{fig:vae}). VAE training was performed without supervision, in an end-to-end encoding-decoding task, optimizing the $\beta$-VAE Variational Lower Bound \citep{higgins2016}, with coefficient $\beta = 0.5$.

In the HPCA training, the learning rate was set to $10^{-3}$. No L2 regularization or dropout was used in this case, since the learning method did not present overfitting issues.

The linear classifiers placed on top of the various network layers were trained with supervision using SGD in the same way as we described above for training the whole network, but the L2 penalty term was reduced to $5 \cdot 10^{-4}$.

To obtain the best possible generalization, \textit{early stopping} was used in each training session, i.e. we chose as final trained model the state of the network at the epoch when the highest validation accuracy was recorded.

All the above mentioned hyperparameters resulted from a parameter search aimed at maximizing the validation accuracy on the respective datasets.

\subsection{Further details on HPCA training}

When HPCA training was used, we noticed that BatchNorm (BN) on higher layers was disruptive. The reason is that HPCA uses the variance along input dimensions in order to understand which dimensions in the input space are more relevant. BN, however, normalizes the input distribution, so that each input dimension is rescaled to have unit variance, thus causing a loss of information that would have been useful to HPCA. For this reason, we defined a modified BN version as follows: instead of dividing each input dimension by the respective variance estimated from samples, we divided each input dimension by the average of all the variances estimated for all the input dimensions. This allowed us to rescale input dimensions in order to have a fixed variance, on average, while at the same time maintaining the relative order of each of the input dimensions in terms of their variances. The modified BN was applied to layers 4 and 5 of the network during HPCA training. On the other hand, standard BN was found to be preferable for earlier layers, where feature detectors had not yet developed a task specificity.

\begin{figure}
\centering
\includegraphics[width=0.4\textwidth]{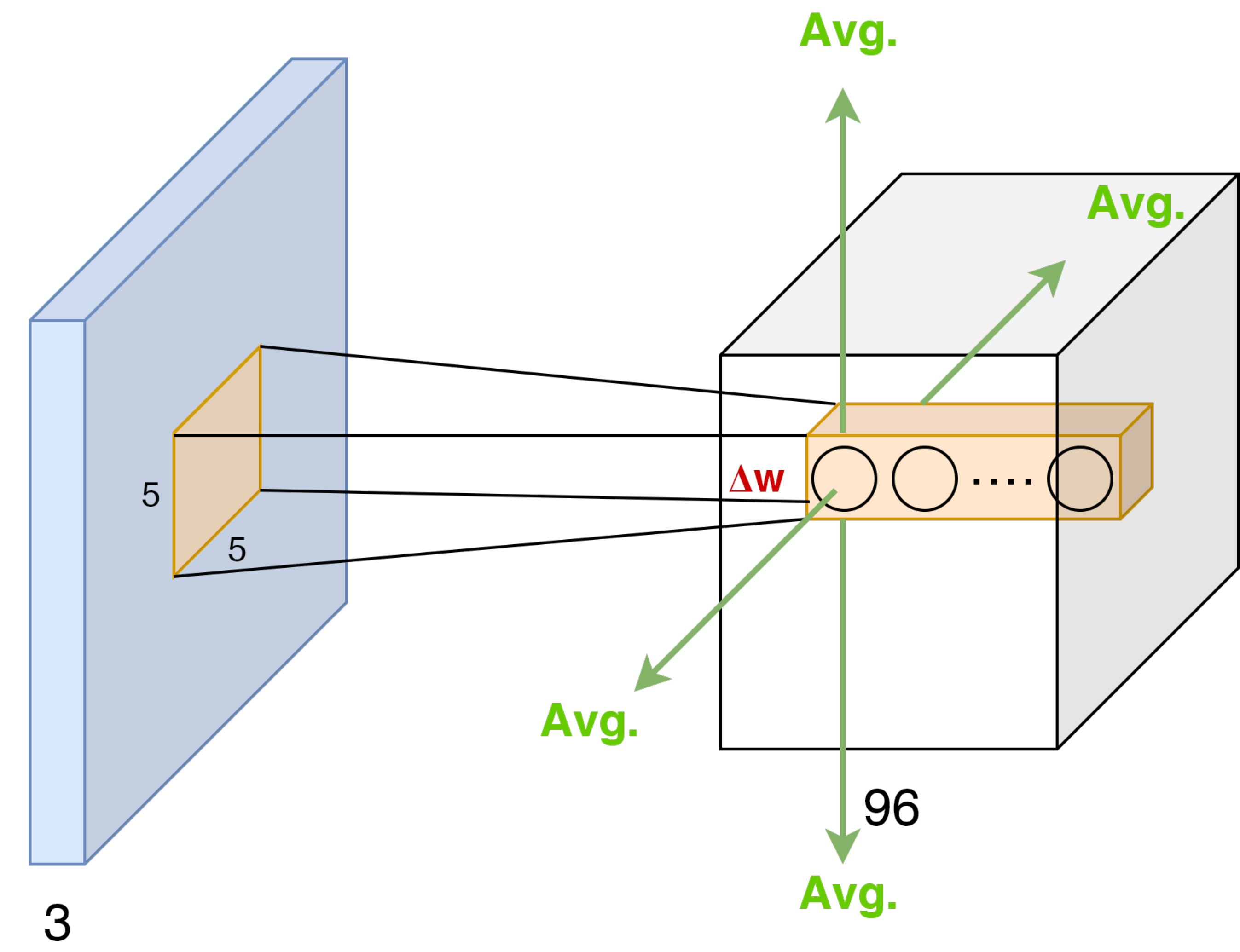}
\caption{Update averaging over horizontal and vertical dimensions.}
\label{fig:conv_layer_upd_avg}
\end{figure}

In convolutional layers, the HPCA rule was applied to convolutional filters at each offset. In order to preserve weight sharing, the resulting weight updates were averaged along the height, width and batch dimensions (Fig. \ref{fig:conv_layer_upd_avg}), and the result was the actual weight update that was applied to the convolutional filter.

\section{Results and discussion} \label{sec:results}

In this section, the experimental results obtained with each dataset are presented and analyzed. Results are reported in Tab. \ref{tab:cifar10_smpleff} Tab. \ref{tab:cifar100_smpleff}, and Tab. \ref{tab:tinyimagenet_smpleff} for CIFAR10, CIFAR100, and Tiny ImageNet, respectively. After training the entire network, independently, with supervised backprop (BP), VAE-based semi-supervised approach (VAE), Hebbian PCA (HPCA), and HPCA plus Fine Tuning (HPCA+FT), we placed and trained a linear classifier on top of the various layers. The linear classifier was trained, on top of the pre-trained networks, using supervised learning with a number of labeled samples corresponding to the various sample efficiency regimes considered. Tables report the classification accuracy, along with the 95\% confidence intervals, for the various sample efficiency regimes, when the classification layer is placed on top of the various internal layers ($L1,\ldots, L5$). 
We report top-1 accuracy for CIFAR10, given that this dataset contains only 10 classes, and top-5 accuracy for CIFAR100 and Tiny ImageNet, given that these datasets contain a much larger number of classes, i.e. 100 and 200, respectively.

\subsection{CIFAR10}

\begin{table*}
    \caption{CIFAR10 accuracy (top-1) and 95\% confidence intervals, obtained with a linear classifier on top of various layers, for the various sample efficiency regimes. Results obtained with supervised backprop (BP), VAE-based semi-supervised approach(VAE), Hebbian PCA (HPCA), and HPCA plus Fine Tuning (HPCA+FT) are compared. It is possible to observe that, in regimes where the number of available samples is low (roughly between 1\% and 5\% of the total available samples), HPCA performs better than BP and VAE approaches in almost all the cases, leading to an improvement up to almost 5\% (on layer 3, in the 1\% regime) w.r.t. non-Hebbian approaches. HPCA+FT helps to further boost accuracy.}
    \begin{center}
        \begin{tabular}{|c|c|c|c|c|c|c|}
            \hline
            \textbf{Regimes} & \textbf{Method} & L1 & L2 & L3 & L4 & L5 \\
            \hline \hline
            \multirow{3}{*}{1\%} 
                & BP   & 33.27 \conf{0.44} & 34.56 \conf{0.34} & 36.80 \conf{0.52} & 35.47 \conf{0.58} & 35.18 \conf{0.57} \\
                \cline{2-7}
                & VAE & 33.54 \conf{0.27} & 34.41 \conf{0.84} & 29.92 \conf{1.25} & 24.91 \conf{0.66} & 22.54 \conf{0.60} \\
                \cline{2-7}
                & HPCA & 36.78 \conf{0.46} & 37.26 \conf{0.14} & 41.31 \conf{0.57} & 39.33 \conf{0.72} & 38.46 \conf{0.44} \\
                \cline{2-7}
                & HPCA+FT & \textbf{37.01} \conf{0.42} & \textbf{37.65} \conf{0.19} & \textbf{41.88} \conf{0.53} & \textbf{40.06} \conf{0.65} & \textbf{39.75} \conf{0.50} \\
            \hline \hline
            \multirow{3}{*}{2\%} 
                & BP   & 37.33 \conf{0.25} & 39.01 \conf{0.19} & 42.34 \conf{0.51} & 41.50 \conf{0.32} & 41.10 \conf{0.39} \\
                \cline{2-7}
                & VAE & 37.65 \conf{0.35} & 39.13 \conf{0.40} & 36.52 \conf{0.47} & 29.39 \conf{0.32} & 26.78 \conf{0.72} \\
                \cline{2-7}
                & HPCA & 41.13 \conf{0.30} & \textbf{41.63} \conf{0.18} & 45.76 \conf{0.41} & 44.70 \conf{0.45} & 43.15 \conf{0.45} \\
                \cline{2-7}
                & HPCA+FT & \textbf{41.60} \conf{0.28} & 42.12 \conf{0.24} & \textbf{46.56} \conf{0.38} & \textbf{45.61} \conf{0.19} & \textbf{45.51} \conf{0.43} \\
            \hline \hline
            \multirow{3}{*}{3\%} 
                & BP   & 40.49 \conf{0.26} & 41.90 \conf{0.40} & 45.13 \conf{0.53} & 45.26 \conf{0.22} & 44.52 \conf{0.24} \\
                \cline{2-7}
                & VAE & 41.22 \conf{0.27} & 43.16 \conf{0.44} & 42.60 \conf{0.87} & 31.91 \conf{0.44} & 29.00 \conf{0.33} \\
                \cline{2-7}
                & HPCA & 44.16 \conf{0.42} & 44.84 \conf{0.08} & 48.92 \conf{0.17} & 47.70 \conf{0.57} & 45.60 \conf{0.27} \\
                \cline{2-7}
                & HPCA+FT & \textbf{44.74} \conf{0.08} & \textbf{45.61} \conf{0.28} & \textbf{49.75} \conf{0.41} & \textbf{48.94} \conf{0.45} & \textbf{48.80} \conf{0.27} \\
            \hline \hline
            \multirow{3}{*}{4\%} 
                & BP   & 43.38 \conf{0.22} & 45.43 \conf{0.18} & 49.51 \conf{0.49} & 48.96 \conf{0.48} & 48.80 \conf{0.24} \\
                \cline{2-7}
                & VAE & 44.39 \conf{0.30} & 45.88 \conf{0.39} & 46.01 \conf{0.40} & 34.26 \conf{0.21} & 31.15 \conf{0.35} \\
                \cline{2-7}
                & HPCA & 46.37 \conf{0.16} & 47.16 \conf{0.28} & 50.70 \conf{0.26} & 49.45 \conf{0.15} & 47.75 \conf{0.54} \\
                \cline{2-7}
                & HPCA+FT & \textbf{47.10} \conf{0.25} & \textbf{48.26} \conf{0.09} & \textbf{52.00} \conf{0.16} & \textbf{51.05} \conf{0.29} & \textbf{51.28} \conf{0.28} \\
            \hline \hline
            \multirow{3}{*}{5\%} 
                & BP   & 45.11 \conf{0.21} & 47.57 \conf{0.29} & 50.61 \conf{0.32} & 50.54 \conf{0.23} & 50.42 \conf{0.14} \\
                \cline{2-7}
                & VAE & 46.31 \conf{0.39} & 48.21 \conf{0.21} & 48.98 \conf{0.34} & 36.32 \conf{0.35} & 32.75 \conf{0.32} \\
                \cline{2-7}
                & HPCA & 47.51 \conf{0.65} & 48.69 \conf{0.37} & 51.69 \conf{0.56} & 50.44 \conf{0.43} & 48.51 \conf{0.32} \\
                \cline{2-7}
                & HPCA+FT & \textbf{48.49} \conf{0.44} & \textbf{50.14} \conf{0.46} & \textbf{53.33} \conf{0.52} & \textbf{52.49} \conf{0.16} & \textbf{52.20} \conf{0.37} \\
            \hline \hline
            \multirow{3}{*}{10\%} 
                & BP   & 51.60 \conf{0.40} & 54.60 \conf{0.31} & 57.97 \conf{0.28} & \textbf{57.63} \conf{0.23} & 57.30 \conf{0.22} \\
                \cline{2-7}
                & VAE & 53.83 \conf{0.26} & \textbf{56.33} \conf{0.22} & 57.85 \conf{0.22} & 52.26 \conf{1.08} & 45.67 \conf{1.15} \\
                \cline{2-7}
                & HPCA & 52.57 \conf{0.29} & 53.29 \conf{0.25} & 56.09 \conf{0.38} & 54.24 \conf{0.28} & 52.68 \conf{0.36} \\
                \cline{2-7}
                & HPCA+FT & \textbf{54.36} \conf{0.32} & 56.08 \conf{0.28} & \textbf{58.46} \conf{0.15} & 56.54 \conf{0.23} & \textbf{57.35} \conf{0.18} \\
            \hline \hline
            \multirow{3}{*}{25\%} 
                & BP   & 60.43 \conf{0.26} & 64.96 \conf{0.18} & 66.63 \conf{0.17} & 68.04 \conf{0.05} & 68.04 \conf{0.20} \\
                \cline{2-7}
                & VAE & \textbf{62.51} \conf{0.24} & \textbf{67.26} \conf{0.32} & \textbf{68.48} \conf{0.21} & \textbf{68.79} \conf{0.29} & \textbf{68.70} \conf{0.15} \\
                \cline{2-7}
                & HPCA & 58.30 \conf{0.28} & 59.20 \conf{0.24} & 59.98 \conf{0.20} & 57.54 \conf{0.20} & 56.46 \conf{0.18} \\
                \cline{2-7}
                & HPCA+FT & 61.45 \conf{0.26} & 65.25 \conf{0.16} & 64.71 \conf{0.17} & 62.43 \conf{0.13} & 64.77 \conf{0.22} \\
            \hline \hline
            \multirow{3}{*}{100\%} 
                & BP   & 61.59 \conf{0.08} & 67.67 \conf{0.11} & 73.87 \conf{0.15} & 83.88 \conf{0.04} & 84.71 \conf{0.02} \\
                \cline{2-7}
                & VAE & \textbf{67.53} \conf{0.22} & \textbf{75.83} \conf{0.31} & \textbf{80.78} \conf{0.28} & \textbf{84.27} \conf{0.35} & \textbf{85.23} \conf{0.26} \\
                \cline{2-7}
                & HPCA & 64.69 \conf{0.29} & 65.92 \conf{0.14} & 64.43 \conf{0.21} & 61.24 \conf{0.22} & 61.16 \conf{0.33} \\
                \cline{2-7}
                & HPCA+FT & 66.76 \conf{0.13} & 75.16 \conf{0.20} & 79.90 \conf{0.18} & 83.55 \conf{0.33} & 84.38 \conf{0.22} \\
            \hline
        \end{tabular}
        \label{tab:cifar10_smpleff}
    \end{center}
\end{table*}

Tab. \ref{tab:cifar10_smpleff} reports the top-1 accuracy results obtained on the CIFAR10 dataset. We only report top-1 accuracy, given that CIFAR10 contains only 10 classes.

At a first glance, we see that in regimes where a limited number of labeled samples is available (between 1\% and 5\%), the HPCA achieves better results than the BP and VAE counterparts, in almost all the cases. On the other hand, when the number of available labeled samples becomes larger, BP and VAE approaches (which exploit end-to-end fine tuning in the supervised phase) are able to take advantage of supervision and improve over HPCA. 
When HPCA+FT is considered, we can observe that end-to-end fine tuning helps to boost accuracy. Still, VAE pre-training performs better in regimes where more labeled samples are available (beyond 10\%), while HPCA and HPCA+FT are preferable in regimes with fewer labeled samples.

Comparing HPCA with the BP approach, we see that, for efficiency regimes up to 3\%, HPCA is better than BP when tested in all layers of the network. We can observe that HPCA generally outperforms backprop by roughly 1-3 percent points, reaching a peak of almost 5 percent points on layer 3, in the 1\% sample efficiency regime. At 4\% efficiency regime, we note that HPCA is still performing better than BP for the first 4 layers, while BP performs better than HPCA when the linear classifier is put on top of the fifth layer. This effect continues for higher efficiency regimes, where we see that increasing the amount of labeled samples, reduces the highest layer where HPCA has better performance than BP. So, we observe that for 5\% efficiency regimes HPCA is better than BP up to Layer 3. Still, in low sample efficiency regimes (between 1\% and 5\%), HPCA outperforms backprop in almost all the cases. At 10\% efficiency regimes HPCA is better only when the linear classifier is trained on top of Layer 1. BP always outperforms HPCA when 100\% labeled examples are used.
To explain this behaviour we can observe that, on one hand, when the amount of labeled samples increases, BP is able to effectively take advantage of the supervised information and extract more useful knowledge from training data. This starts to be seen from the highest layers of the network, where the supervision signal is stronger. Increasing the amount of labeled training data brings this effect down up to the first network layer.
On the other hand, unsupervised Hebbian learning signal, which is driven by the inputs, is stronger in the first layers, where coherence between input and output of neurons is more meaningful, and layers tend to adapt faster to the unsupervised stimuli.

Comparing HPCA with VAE approach, we see that, when low sample efficiency regimes are considered (between 1\% and 5\%) Hebbian approaches always achieve significantly higher results than VAE. Only when the number of available labeled samples increases (beyond 10\%), VAE pre-training starts to become really competitive, obtaining results comparable to or higher than HPCA. In these scenarios, VAE pre-training also helps improving performance w.r.t. plain BP training from scratch. 
We can also observe that, in low sample efficiency regimes (10 \% or less), the VAE approach suffers from a decrease in performance when going deeper with the number of layers. This issue is common with unsupervised methods, because the lack of a supervision signal (or still its scarcity, in case of semi-supervised scenarios) makes it more difficult to develop task-specific feature detectors on higher layers, which are essential to reach higher performances, as also previous studies on deep CNNs reveal \citep{agrawal2014}. With HPCA, this problem seems to alleviate, and the accuracy remains pretty much constant with the number of layers, meaning that the features produced by this approach are more meaningful for the classification task. Only when the amount of supervision, i.e. the number of labeled samples, becomes large enough (above 10\%), the end-to-end supervised training phase, that follows VAE pre-training in the semi-supervised approach, manages to transform VAE feature detectors to task-specific features that perform even better than those obtained by BP training from scratch.
Overall, these results suggest that VAE-based semi-supervised learning is better suited in sample efficiency regimes where the labeled portion of the dataset is still relatively large (10\% or more), while our method is preferable to address sample efficiency regimes in which the number of available labeled samples is \textit{very} small (5\% or less).

The HPCA+FT strategy is still preferable in low sample efficiency regimes (between 1\% and 5\%), where it helps to further increase accuracy w.r.t. plain HPCA. In particular, in these regimes, we can observe a further increase in accuracy up to 2\% points on network layer 3 and 4, and up to 4\% points on layer 5 (in the 4\% and 5\% regimes). Fine tuning also helps increasing accuracy in successive sample efficiency regimes, especially on higher layers.

Please note that in some configurations, the accuracy of pure HPCA (even without fine-tuning on deep layers) is higher than the accuracy obtained with BP. Consider, for instance, the accuracy obtained at L5 for configurations up to 3\% sample efficiency regimes. This might appear strange since, in principle, label information used by BP should help to achieve higher accuracy. The explanation is that BP training tends to generalize poorly with low sample efficiency regimes, when just a limited number of labeled samples is available. On the other hand, unsupervised HPCA training exploits a large number of unlabeled samples, and this allows us to achieve generally higher accuracy in the aforementioned cases.

\subsection{CIFAR100}

\begin{table*}
    \caption{CIFAR100 accuracy (top-5) and 95\% confidence intervals obtained with a linear classifier on top of various layers, for the various sample efficiency regimes. Results obtained with supervised backprop (BP), VAE-based semi-supervised approach (VAE), Hebbian PCA (HPCA), and HPCA plus Fine Tuning (HPCA+FT) are compared. It is possible to observe that, in regimes where the number of available samples is low (roughly between 1\% and 5\% of the total available samples), HPCA performs better than BP and VAE approaches in almost all the cases. Even though, in various cases, the improvement is small, it becomes significant in some scenarios, where peaks of improvement up to 2\% are observed (on layers 3 and 4) w.r.t. non-Hebbian approaches. HPCA+FT helps to further boost accuracy.}
    \begin{center}
        \begin{tabular}{|c|c|c|c|c|c|c|}
            \hline
            \textbf{Regimes} & \textbf{Method} & L1 & L2 & L3 & L4 & L5 \\
            \hline \hline
            \multirow{3}{*}{1\%} 
                & BP   & \textbf{22.56} \conf{0.53} & \textbf{22.73} \conf{0.28} & 23.41 \conf{0.44} & 20.85 \conf{0.58} & 21.88 \conf{0.30} \\
                \cline{2-7}
                & VAE & 21.69 \conf{0.10} & 21.70 \conf{0.30} & 17.61 \conf{0.54} & 13.45 \conf{0.54} & 12.28 \conf{0.50} \\
                \cline{2-7}
                & HPCA & 21.90 \conf{0.33} & 22.23 \conf{0.50} & 22.98 \conf{0.18} & 20.88 \conf{0.43} & 21.90 \conf{0.55} \\
                \cline{2-7}
                & HPCA+FT & 22.30 \conf{0.38} & 22.28 \conf{0.63} & \textbf{23.58} \conf{0.21} & \textbf{21.70} \conf{0.61} & \textbf{22.63} \conf{0.55} \\
            \hline \hline
            \multirow{3}{*}{2\%} 
                & BP   & 28.99 \conf{0.49} & 29.07 \conf{0.68} & 30.75 \conf{0.34} & 27.67 \conf{0.37} & 28.18 \conf{0.35} \\
                \cline{2-7}
                & VAE & 28.24 \conf{0.13} & 28.42 \conf{0.31} & 23.56 \conf{0.73} & 17.01 \conf{0.37} & 15.25 \conf{0.63} \\
                \cline{2-7}
                & HPCA & 29.08 \conf{0.31} & \textbf{29.40} \conf{0.23} & 32.22 \conf{0.28} & 29.20 \conf{0.46} & 28.95 \conf{0.35} \\
                \cline{2-7}
                & HPCA+FT & \textbf{29.65} \conf{0.52} & 26.57 \conf{0.26} & \textbf{33.20} \conf{0.20} & \textbf{30.21} \conf{0.54} & \textbf{30.83} \conf{0.35} \\
            \hline \hline
            \multirow{3}{*}{3\%} 
                & BP   & 31.77 \conf{0.42} & 32.56 \conf{0.51} & 34.06 \conf{0.41} & 31.81 \conf{0.33} & 32.45 \conf{0.23} \\
                \cline{2-7}
                & VAE & 31.28 \conf{0.54} & 31.71 \conf{0.27} & 27.46 \conf{1.23} & 18.26 \conf{0.24} & 16.44 \conf{0.12} \\
                \cline{2-7}
                & HPCA & 32.07 \conf{0.46} & 33.04 \conf{0.30} & 36.41 \conf{0.15} & 33.67 \conf{0.39} & 32.61 \conf{0.51} \\
                \cline{2-7}
                & HPCA+FT & \textbf{32.81} \conf{0.18} & \textbf{33.08} \conf{0.55} & \textbf{37.75} \conf{0.38} & \textbf{35.02} \conf{0.36} & \textbf{35.04} \conf{0.17} \\
            \hline \hline
            \multirow{3}{*}{4\%} 
                & BP   & 34.74 \conf{0.29} & 35.88 \conf{0.30} & 37.63 \conf{0.19} & 35.92 \conf{0.35} & 36.52 \conf{0.37} \\
                \cline{2-7}
                & VAE & 34.60 \conf{0.10} & 35.44 \conf{0.31} & 32.34 \conf{0.79} & 19.68 \conf{0.32} & 17.89 \conf{0.27} \\
                \cline{2-7}
                & HPCA & 35.34 \conf{0.40} & 35.97 \conf{0.27} & 39.85 \conf{0.35} & 37.23 \conf{0.19} & 36.05 \conf{0.37} \\
                \cline{2-7}
                & HPCA+FT & \textbf{36.13} \conf{0.39} & \textbf{36.23} \conf{0.20} & \textbf{41.21} \conf{0.39} & \textbf{39.16} \conf{0.32} & \textbf{38.89} \conf{0.15} \\
            \hline \hline
            \multirow{3}{*}{5\%} 
                & BP   & 36.84 \conf{0.23} & 37.70 \conf{0.32} & 39.70 \conf{0.21} & 38.42 \conf{0.32} & 39.21 \conf{0.65} \\
                \cline{2-7}
                & VAE & 36.68 \conf{0.17} & 37.26 \conf{0.26} & 35.33 \conf{0.81} & 20.55 \conf{0.44} & 18.48 \conf{0.26} \\
                \cline{2-7}
                & HPCA & 37.28 \conf{0.40} & 37.75 \conf{0.24} & 42.12 \conf{0.49} & 39.37 \conf{0.18} & 37.84 \conf{0.22} \\
                \cline{2-7}
                & HPCA+FT & \textbf{38.03} \conf{0.20} & \textbf{38.02} \conf{0.25} & \textbf{43.76} \conf{0.33} & \textbf{41.66} \conf{0.20} & \textbf{41.42} \conf{0.23} \\
            \hline \hline
            \multirow{3}{*}{10\%} 
                & BP   & 42.04 \conf{0.24} & \textbf{44.98} \conf{0.23} & 48.39 \conf{0.22} & 48.98 \conf{0.35} & \textbf{49.84} \conf{0.34} \\
                \cline{2-7}
                & VAE & 42.64 \conf{0.34} & 44.84 \conf{0.48} & 46.04 \conf{0.44} & 27.81 \conf{0.13} & 23.80 \conf{0.60} \\
                \cline{2-7}
                & HPCA & 43.05 \conf{0.36} & 43.93 \conf{0.23} & 48.68 \conf{0.27} & 46.05 \conf{0.24} & 43.87 \conf{0.28} \\
                \cline{2-7}
                & HPCA+FT & \textbf{43.51} \conf{0.34} & 44.84 \conf{0.26} & \textbf{50.84} \conf{0.22} & \textbf{49.53} \conf{0.19} & 48.93 \conf{0.38} \\
            \hline \hline
            \multirow{3}{*}{25\%} 
                & BP   & 53.36 \conf{0.10} & \textbf{59.11} \conf{0.21} & 60.94 \conf{0.15} & \textbf{64.57} \conf{0.26} & \textbf{67.17} \conf{0.16} \\
                \cline{2-7}
                & VAE & \textbf{53.53} \conf{0.12} & 57.63 \conf{0.52} & \textbf{62.16} \conf{0.57} & 55.29 \conf{0.68} & 52.59 \conf{1.02} \\
                \cline{2-7}
                & HPCA & 49.62 \conf{0.36} & 51.30 \conf{0.25} & 56.14 \conf{0.29} & 53.46 \conf{0.28} & 51.29 \conf{0.15} \\
                \cline{2-7}
                & HPCA+FT & 51.51 \conf{0.31} & 54.22 \conf{0.23} & 59.60 \conf{0.44} & 58.29 \conf{0.29} & 58.70 \conf{0.18} \\
            \hline \hline
            \multirow{3}{*}{100\%} 
                & BP   & 51.67 \conf{0.10} & 60.84 \conf{0.19} & 67.01 \conf{0.13} & 78.85 \conf{0.10} & \textbf{80.74} \conf{0.05} \\
                \cline{2-7}
                & VAE & \textbf{67.51} \conf{0.11} & \textbf{73.83} \conf{0.30} & \textbf{78.70} \conf{0.23} & \textbf{79.58} \conf{0.18} & 79.97 \conf{0.14} \\
                \cline{2-7}
                & HPCA & 60.94 \conf{0.09} & 62.24 \conf{0.15} & 64.17 \conf{0.22} & 61.27 \conf{0.24} & 59.51 \conf{0.20} \\
                \cline{2-7}
                & HPCA+FT & 65.61 \conf{0.12} & 70.38 \conf{0.23} & 74.10 \conf{0.12} & 73.38 \conf{0.18} & 74.42 \conf{0.14} \\
            \hline
        \end{tabular}
        \label{tab:cifar100_smpleff}
    \end{center}
\end{table*}

Since CIFAR10 contained just 10 different classes, to validate our observations with a similar, yet more difficult scenario, we also performed tests with CIFAR100, containing 100 classes.
In Tab. \ref{tab:cifar100_smpleff} the top-5 accuracy results obtained on the CIFAR100 dataset are shown. 
In this case, we report top-5 accuracy instead of top-1, given that CIFAR100 contains a much larger number of classes than the previous dataset.

These experiments confirm our previous observations. The results show that, in regimes where a limited number of labeled samples is available (between 1\% and 5\%), our semi-supervised approach, based on Hebbian learning, achieves better results than BP and VAE counterparts in almost all the cases. On the other hand, when the number of available labeled samples becomes larger, BP and VAE approaches (which exploit end-to-end fine tuning in the supervised phase) are able to take advantage of supervision and improve over HPCA.
Also in this case, we can observe that the end-to-end fine tuning in HPCA+FT helps to further boost accuracy.

Except for 1\% sample efficiency regime, where the difference in the results is not really significant, for regimes up to 3\% HPCA is always better than BP. In particular, the improvement becomes significant in correspondence of network layers 3 and 4 in the sample efficiency regimes between 2\% and 5\%. Note also that layer 3 generally offers absolute highest accuracy for all efficiency regimes, when HPCA is used. In these cases, we observe peaks of improvement over 2 percent points. As before, for efficiency regimes higher than 4\% BP starts to provide better accuracy in the higher layers of the network. However, HPCA is still better than BP, with 100\% efficiency regime, when tested on Layer 1 and 2. This can be explained by the fact that CIFAR100 offers a more difficult scenario than CIFAR10, and BP has more problems than before in backpropagating the error signal to the initial layers of the network. Differently, unsupervised Hebbian learning has most of its effects in the very first layers of the network.


Also in this case, we observe that HPCA always performs better than VAE method when low sample efficiency regimes are considered (between 1\% and 5\%), especially for higher network layers. Again, VAE pre-training seems to be more effective in regimes where more labeled samples are available (beyond 10\%).

The HPCA+FT strategy is still preferable in low sample efficiency regimes (between 1\% and 5\%), where it helps to further increase accuracy w.r.t. plain HPCA. In particular, in these regimes, we can observe a further increase in accuracy up to 4\% points on layer 5 (in the 5\% regime). Fine tuning also helps increasing accuracy in successive sample efficiency regimes, especially on higher layers.

\subsection{Tiny ImageNet}

\begin{table*}
    \caption{Tiny ImageNet accuracy (top-5) and 95\% confidence intervals obtained with a linear classifier on top of various layers, for the various sample efficiency regimes. Results obtained with supervised backprop (BP), VAE-based semi-supervised approach (VAE), Hebbian PCA (HPCA), and HPCA plus Fine Tuning (HPCA+FT) are compared. It is possible to observe that, in regimes where the number of available samples is low (roughly between 1\% and 5\% of the total available samples), HPCA performs better than BP and VAE approaches in almost all the cases, leading to an improvement up to almost 3\% (on layer 3, in the 4\% regime)  w.r.t. non-Hebbian approaches. HPCA+FT helps to further boost accuracy.
    }
    \begin{center}
        \begin{tabular}{|c|c|c|c|c|c|c|}
            \hline
            \textbf{Regimes} & \textbf{Method} & L1 & L2 & L3 & L4 & L5 \\
            \hline \hline
            \multirow{3}{*}{1\%} 
                & BP   & 9.89 \conf{0.15} & 10.10 \conf{0.26} & 9.99 \conf{0.17} & 9.15 \conf{0.23} & 9.53 \conf{0.21} \\
                \cline{2-7}
                & VAE & 9.63 \conf{0.26} & 9.49 \conf{0.39} & 7.58 \conf{0.28} & 5.99 \conf{0.19} & 5.55 \conf{0.23} \\
                \cline{2-7}
                & HPCA & \textbf{10.83} \conf{0.28} & 10.87 \conf{0.26} & 11.85 \conf{0.19} & 10.84 \conf{0.26} & 10.86 \conf{0.23} \\
                \cline{2-7}
                & HPCA+FT & 10.81 \conf{0.27} & \textbf{10.99} \conf{0.36} & \textbf{12.15} \conf{0.46} & \textbf{11.05} \conf{0.27} & \textbf{11.38} \conf{0.41} \\
            \hline \hline
            \multirow{3}{*}{2\%} 
                & BP   & 12.76 \conf{0.27} & 12.84 \conf{0.14} & 13.95 \conf{0.34} & 13.04 \conf{0.15} & 13.48 \conf{0.39} \\
                \cline{2-7}
                & VAE & 12.94 \conf{0.37} & 13.06 \conf{0.23} & 10.86 \conf{0.28} & 7.40 \conf{0.27} & 6.74 \conf{0.20} \\
                \cline{2-7}
                & HPCA & 13.84 \conf{0.17} & \textbf{14.35} \conf{0.15} & 16.18 \conf{0.15} & 14.52 \conf{0.32} & 14.03 \conf{0.15} \\
                \cline{2-7}
                & HPCA+FT & \textbf{14.12} \conf{0.23} & 14.32 \conf{0.31} & \textbf{16.89} \conf{0.61} & \textbf{15.28} \conf{0.28} & \textbf{15.71} \conf{0.47} \\
            \hline \hline
            \multirow{3}{*}{3\%} 
                & BP   & 14.12 \conf{0.20} & 14.65 \conf{0.57} & 16.50 \conf{0.32} & 15.76 \conf{0.27} & 15.99 \conf{0.38} \\
                \cline{2-7}
                & VAE & 14.31 \conf{0.18} & 15.17 \conf{0.20} & 13.67 \conf{0.36} & 8.35 \conf{0.29} & 7.74 \conf{0.19} \\
                \cline{2-7}
                & HPCA & 16.13 \conf{0.14} & 16.32 \conf{0.33} & 18.87 \conf{0.29} & 17.04 \conf{0.26} & 16.38 \conf{0.25} \\
                \cline{2-7}
                & HPCA+FT & \textbf{16.25} \conf{0.21} & \textbf{16.54} \conf{0.28} & \textbf{19.78} \conf{0.47} & \textbf{18.31} \conf{0.24} & \textbf{18.23} \conf{0.33} \\
            \hline \hline
            \multirow{3}{*}{4\%} 
                & BP   & 15.44 \conf{0.42} & 16.72 \conf{0.31} & 18.36 \conf{0.22} & 17.85 \conf{0.16} & 17.84 \conf{0.19} \\
                \cline{2-7}
                & VAE & 16.09 \conf{0.20} & 17.05 \conf{0.20} & 16.83 \conf{0.51} & 8.86 \conf{0.11} & 8.45 \conf{0.21} \\
                \cline{2-7}
                & HPCA & 17.64 \conf{0.49} & 18.27 \conf{0.34} & 21.07 \conf{0.17} & 19.16 \conf{0.33} & 18.13 \conf{0.39} \\
                \cline{2-7}
                & HPCA+FT & \textbf{17.70} \conf{0.44} & \textbf{18.33} \conf{0.24} & \textbf{21.95} \conf{0.57} & \textbf{20.86} \conf{0.32} & \textbf{20.55} \conf{0.28} \\
            \hline \hline
            \multirow{3}{*}{5\%} 
                & BP   & 16.75 \conf{0.25} & 17.94 \conf{0.25} & 20.26 \conf{0.21} & 20.15 \conf{0.35} & 19.84 \conf{0.36} \\
                \cline{2-7}
                & VAE & 17.44 \conf{0.26} & 18.62 \conf{0.32} & 19.16 \conf{0.52} & 9.92 \conf{0.24} & 9.29 \conf{0.17} \\
                \cline{2-7}
                & HPCA & 18.93 \conf{0.14} & 19.67 \conf{0.36} & 22.65 \conf{0.35} & 21.01 \conf{0.38} & 19.57 \conf{0.15} \\
                \cline{2-7}
                & HPCA+FT & \textbf{19.26} \conf{0.41} & \textbf{19.93} \conf{0.41} & \textbf{23.97} \conf{0.52} & \textbf{22.95} \conf{0.26} & \textbf{22.46} \conf{0.17} \\
            \hline \hline
            \multirow{3}{*}{10\%} 
                & BP   & 20.26 \conf{0.18} & 23.12 \conf{0.14} & 27.05 \conf{0.20} & 27.30 \conf{0.20} & 27.21 \conf{0.29} \\
                \cline{2-7}
                & VAE & 21.62 \conf{0.25} & 23.83 \conf{0.19} & 27.42 \conf{0.18} & 16.69 \conf{0.18} & 13.51 \conf{0.34} \\
                \cline{2-7}
                & HPCA & 22.15 \conf{0.43} & 23.69 \conf{0.24} & 27.02 \conf{0.24} & 25.73 \conf{0.34} & 23.08 \conf{0.17} \\
                \cline{2-7}
                & HPCA+FT & \textbf{22.82} \conf{0.33} & \textbf{24.34} \conf{0.29} & \textbf{28.69} \conf{0.36} & \textbf{28.79} \conf{0.26} & \textbf{28.13} \conf{0.38} \\
            \hline \hline
            \multirow{3}{*}{25\%} 
                & BP   & 28.97 \conf{0.26} & \textbf{32.63} \conf{0.36} & 37.38 \conf{0.13} & \textbf{38.81} \conf{0.20} & \textbf{38.80} \conf{0.39} \\
                \cline{2-7}
                & VAE & \textbf{29.40} \conf{0.31} & 32.42 \conf{0.29} & \textbf{39.93} \conf{0.31} & 37.97 \conf{0.62} & 37.89 \conf{0.54} \\
                \cline{2-7}
                & HPCA & 27.05 \conf{0.47} & 28.39 \conf{0.34} & 32.08 \conf{0.19} & 31.30 \conf{0.26} & 29.51 \conf{0.23} \\
                \cline{2-7}
                & HPCA+FT & 28.01 \conf{0.75} & 30.63 \conf{0.16} & 35.87 \conf{0.53} & 36.98 \conf{0.26} & 37.10 \conf{0.23} \\
            \hline \hline
            \multirow{3}{*}{100\%} 
                & BP   & \textbf{42.89} \conf{0.13} & \textbf{49.94} \conf{0.13} & 54.54 \conf{0.27} & 57.00 \conf{0.16} & 57.50 \conf{0.16} \\
                \cline{2-7}
                & VAE & 42.32 \conf{0.16} & 48.54 \conf{0.53} & \textbf{58.31} \conf{0.12} & \textbf{59.60} \conf{0.23} & \textbf{60.23} \conf{0.65} \\
                \cline{2-7}
                & HPCA & 35.74 \conf{0.15} & 38.29 \conf{0.19} & 38.78 \conf{0.07} & 38.61 \conf{0.21} & 36.99 \conf{0.36} \\
                \cline{2-7}
                & HPCA+FT & 40.34 \conf{0.31} & 45.00 \conf{0.40} & 53.12 \conf{0.26} & 52.95 \conf{0.28} & 53.96 \conf{0.43} \\
            \hline
        \end{tabular}
        \label{tab:tinyimagenet_smpleff}
    \end{center}
\end{table*}

Further experiments on Tiny ImageNet allowed us to validate the scalability of our previous observations to larger datasets. Tiny ImageNet has 200 classes and the training set consists of 100,000 samples (90,000 of which are used for training and 10,000 for validation).
Results are reported in Tab. \ref{tab:tinyimagenet_smpleff}, were the top-5 accuracy measures are shown, along with their 95\% confidence interval.
Also in this case, we report top-5 accuracy instead of top-1, given that Tiny ImageNet contains a large number of classes, as opposed to other datasets such as CIFAR10.

Again, results confirm our observations. In regimes where a limited number of labeled samples is available (between 1\% and 5\%), the Hebbian approach outperforms BP and VAE counterparts, in almost all the cases. On the other hand, when the number of available labeled samples becomes larger, BP and VAE approaches (which exploit end-to-end fine tuning in the supervised phase) are able to take advantage of supervision and improve over HPCA. 
Also in this case, we can observe that the end-to-end fine tuning in HPCA+FT helps to further boost accuracy.

Specifically, HPCA outperforms BP in all layers up to 4\% sample efficiency regime. In fact, with CIFAR10 and CIFAR100, BP started to outperform HPCA on layer 5 at 4\% regimes. Here, with 4\% sample efficiency regime, HPCA is still better than BP in all layers. This is probably due to the fact that the number of classes in Tiny ImageNet is higher and using just a few samples does not allow back propagation to correctly adapt the behaviour of network layers. In addition, we can observe that HPCA generally outperforms backprop by roughly 1-2 percent points, reaching a peak of almost 3 percent points on layer 3, in the 4\% sample efficiency regime.
With higher efficiency regimes, BP begins to outperform HPCA, starting from the higher layers. At 100\% sample efficiency regime, BP outperforms HPCA on all layers. This is probably due to the fact that 90,000 labeled training samples are sufficient for BP to correctly train all network layers, exploiting the supervised information.

Also in this case, we observe that HPCA always performs better than VAE method when low sample efficiency regimes are considered (between 1\% and 5\%), especially for higher network layers. Again, VAE pre-training seems to be more effective in regimes where more labeled samples are available (beyond 10\%).

The HPCA+FT strategy is still preferable in low sample efficiency regimes (between 1\% and 5\%), where it helps to further increase accuracy w.r.t. plain HPCA. In particular, in these regimes, we can observe a further increase in accuracy up to 3\% points on layer 5 (in the 5\% regime). Fine tuning also helps increasing accuracy in successive sample efficiency regimes, especially on higher layers.

\subsection{Effect of pooling layers}

\begin{figure*}
    \centering
    \subfloat[Accuracy results (top-1) in the 1\% sample efficiency regime.]{
        \includegraphics[width=0.5\textwidth]{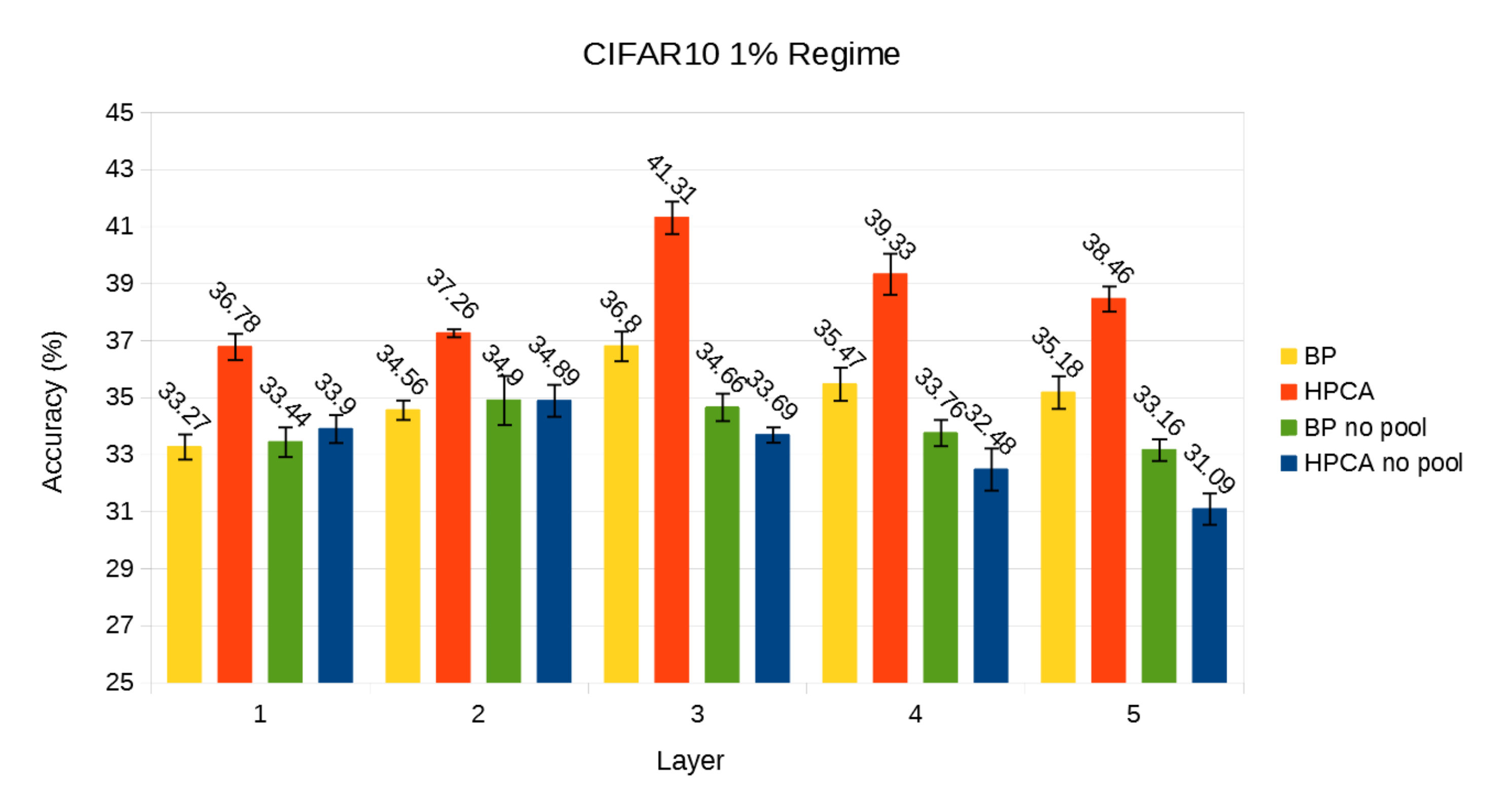}
        \label{fig:cifar10_smpleff_pool_1}
    }
    ~
    \subfloat[Accuracy results (top-1) in the 5\% sample efficiency regime.]{
        \includegraphics[width=0.5\textwidth]{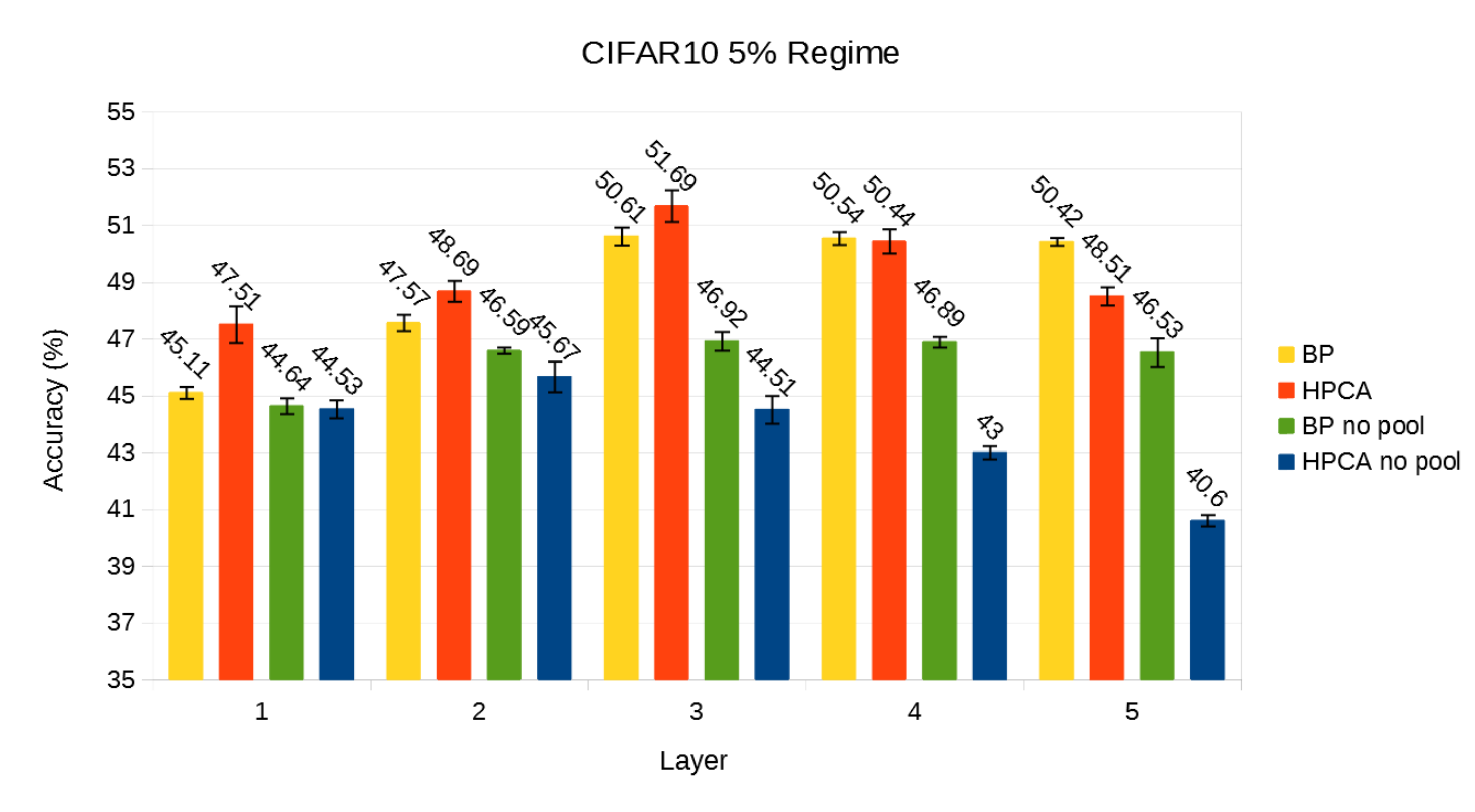}
    \label{fig:cifar10_smpleff_pool_5}
    }
    \caption{Comparison of network architectures with and without pooling on the CIFAR10 dataset.}
    \label{fig:cifar10_smpleff_no_pool}
\end{figure*}

\begin{figure*}
    \centering
    \subfloat[Accuracy results (top-5) in the 1\% sample efficiency regime.]{
        \includegraphics[width=0.5\textwidth]{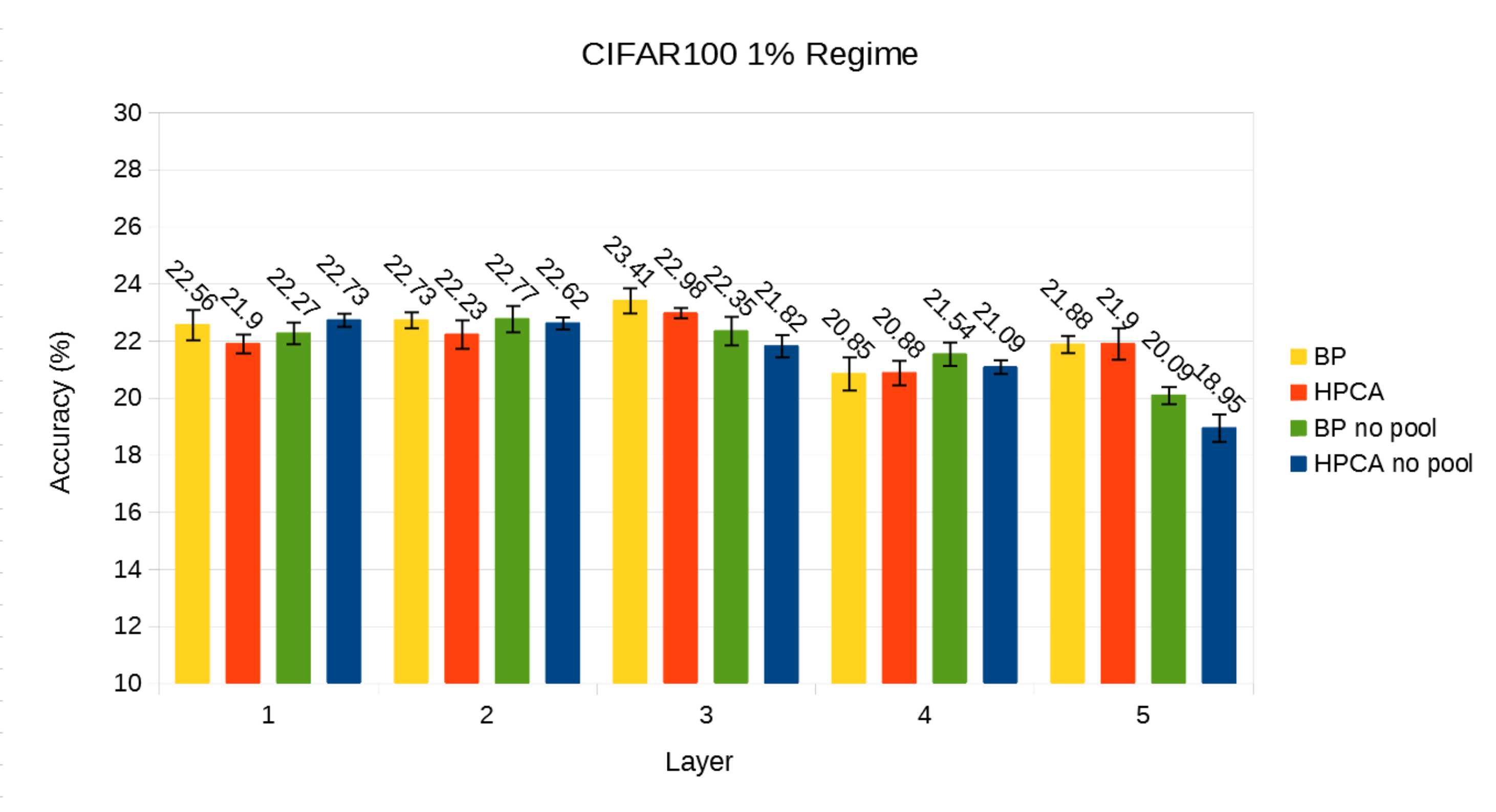}
        \label{fig:cifar100_smpleff_pool_1}
    }
    ~
    \subfloat[Accuracy results (top-5) in the 5\% sample efficiency regime.]{
        \includegraphics[width=0.5\textwidth]{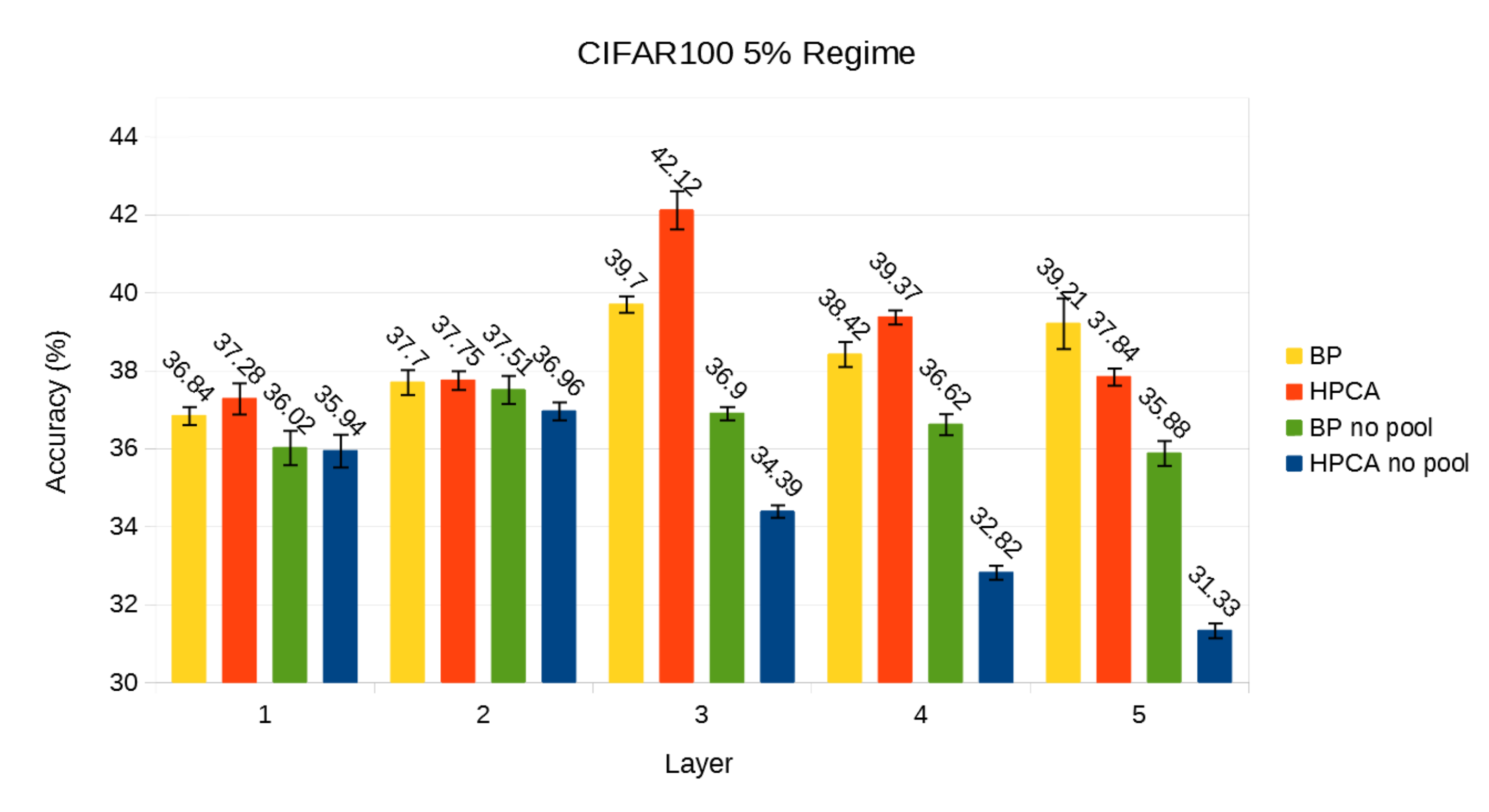}
    \label{fig:cifar100_smpleff_pool_5}
    }
    \caption{Comparison of network architectures with and without pooling on the CIFAR100 dataset.}
    \label{fig:cifar100_smpleff_no_pool}
\end{figure*}

\begin{figure*}
    \centering
    \subfloat[Accuracy results (top-5) in the 1\% sample efficiency regime.]{
        \includegraphics[width=0.5\textwidth]{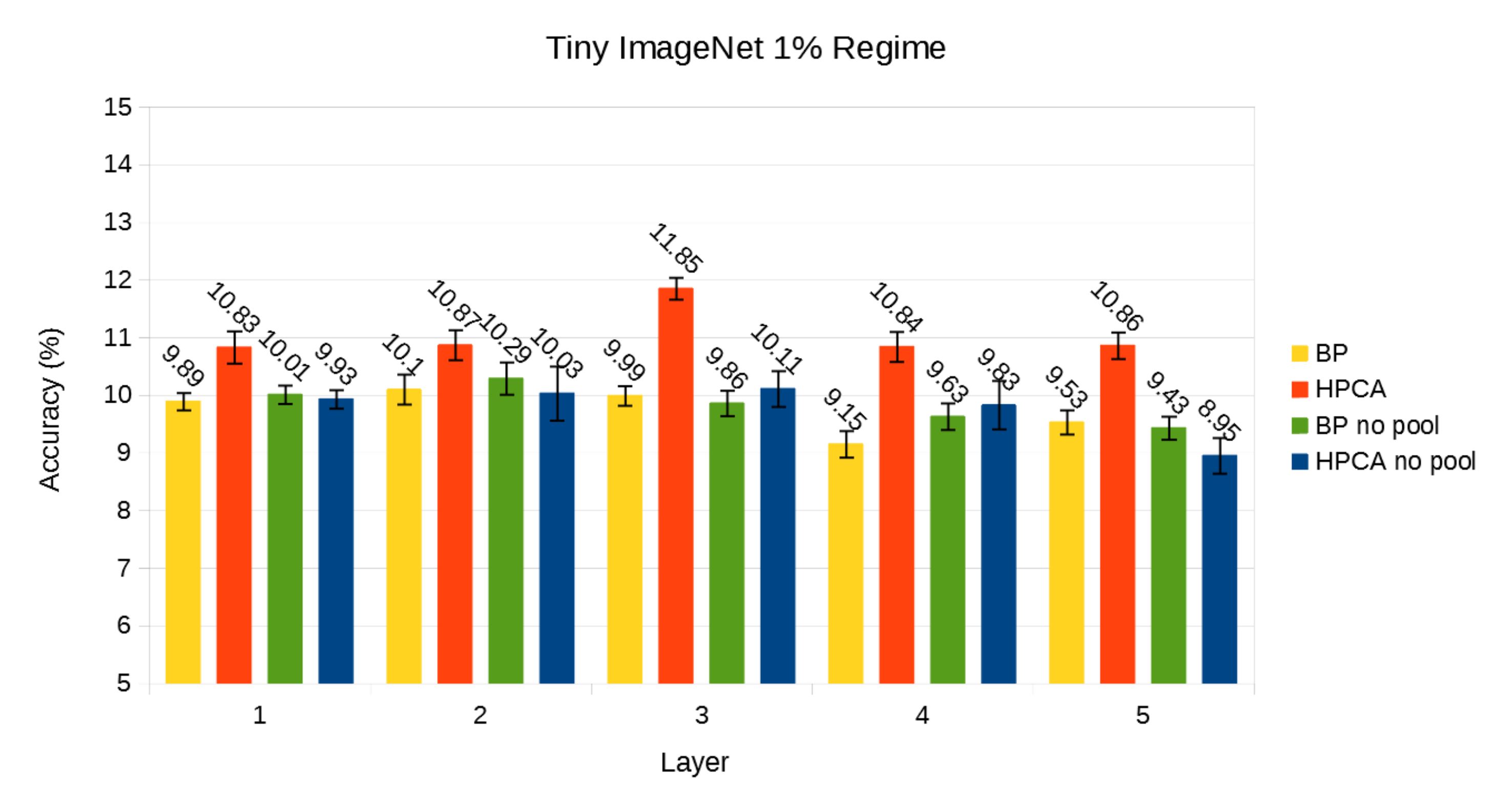}
        \label{fig:tinyimagenet_smpleff_pool_1}
    }
    ~
    \subfloat[Accuracy results (top-5) in the 5\% sample efficiency regime.]{
        \includegraphics[width=0.5\textwidth]{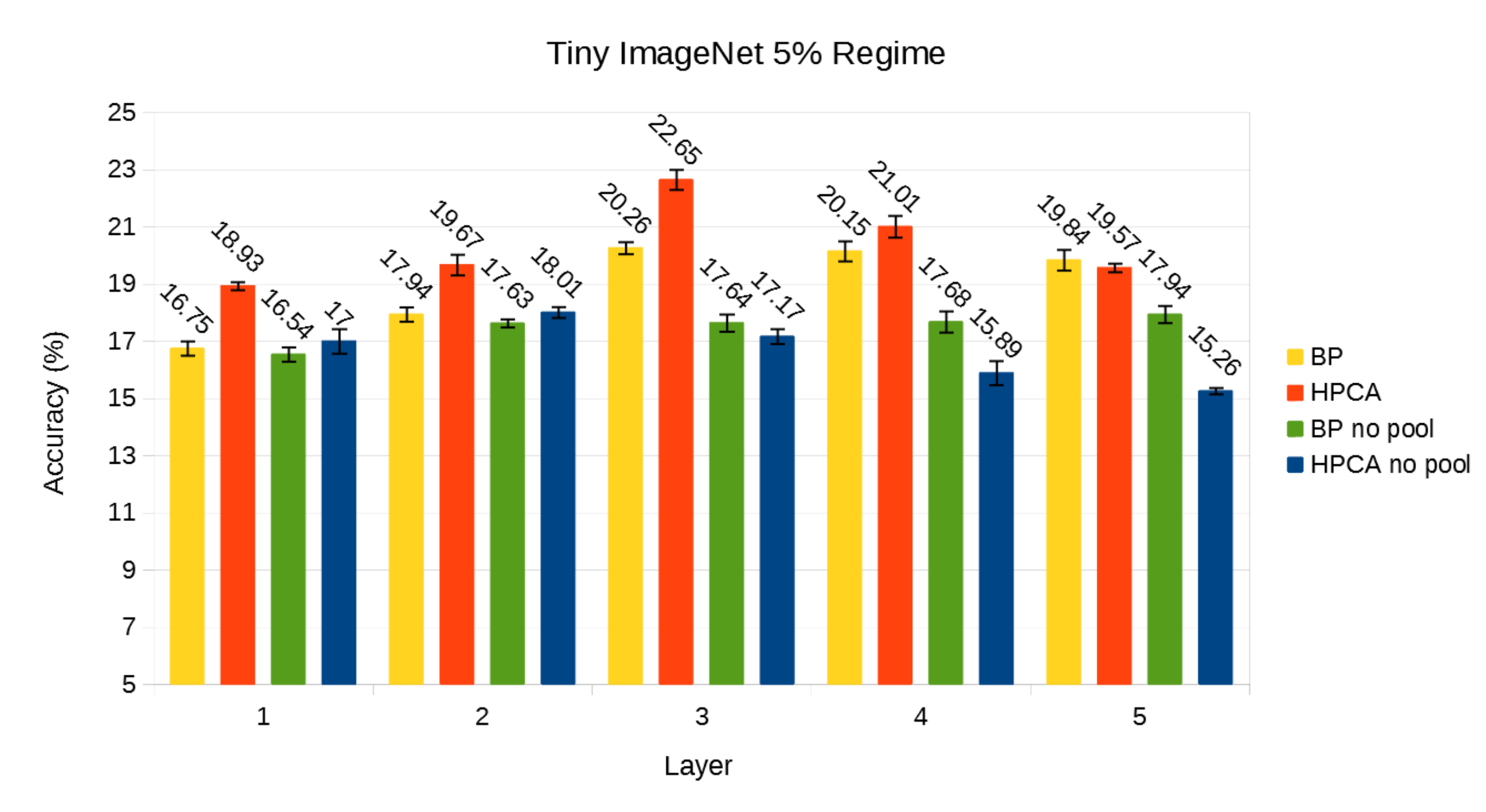}
    \label{fig:tinyimagenet_smpleff_pool_5}
    }
    \caption{Comparison of network architectures with and without pooling on the Tiny ImageNet dataset.}
    \label{fig:tinyimagenet_smpleff_no_pool}
\end{figure*}

From the results presented so far, it is possible to observe that higher results for low sample efficiency regimes (1-5\%) are generally achieved in correspondence of layer 3. We note also that in Layer 1 and 3 of our network, as show in Fig. \ref{fig:network}, we have max-pooling operations. We performed further experiments in order to evaluate the impact of pooling layers on the final results. In this subsection, we discuss the results of previous experiments executed in a network where all max-pooling operations were eliminated. 
In Figs. \ref{fig:cifar10_smpleff_no_pool}, \ref{fig:cifar100_smpleff_no_pool}, \ref{fig:tinyimagenet_smpleff_no_pool}, we show the accuracy obtained for the 1\% and 5\% sample efficiency regimes, on the CIFAR10, CIFAR100 and Tiny ImageNet datasets.
Yellow and red bars correspond to experiments executed with the original network, trained with BP and HPCA, respectively. Green and blue bars corresponds to experiments executed with the network where max-pooling was eliminated, also trained with BP and HPCA respectively. 
We show the results for BP and HPCA, because these are the scenarios in which the aforementioned effect is more prominent.

It can be clearly seen the peak of accuracy, occurring at layer 3, for all experiments executed with the original network. Similarly, we can see that the peak disappeared with the experiments executed without max-pooling. In fact, it seems that the peak moved at layer 2. However, it is within the reported confidence interval of the adjacent layers, so the difference is not statistically significant. In addition, when max-pooling is not used, we report a general decrease of performance. This confirms that max-pooling is a relevant element and its use significantly helps our semi-supervised approach. Note that also in case of BP experiments, without max-pooling, significantly lower results were obtained w.r.t. the highest accuracy, which was achieved with HPCA training in conjunction with pooling layers.

Max-pooling clearly helps the network to produce better feature maps to be used by the linear classifier. In fact, adding consecutive convolutional layers produces neurons with increasingly larger receptive field size. However neurons activations corresponding to adjacent areas, in a given feature map, will be highly correlated. The effect of pooling is to reduce this correlation. This turns out to be very helpful for the final classifiers, which can better handle feature maps of lower dimension and with less redundant information, making it easier to discover relationships between features and target classes.

It can also be observed that, when the pooling layer is removed, the accuracy drop of HPCA is larger than that of BP. Again this can be justified by considering that, without pooling, neuron activations corresponding to adjacent areas will be highly correlated. Nonetheless, when backprop training is used, the supervision signal can drive network weights in order to produce feature representations that reduce such correlation. With unsupervised Hebbian training, this is not possible, and therefore the resulting model suffers a higher performance drop when correlations are incentivized due to the removal of the pooling layers. So our conclusion is that pooling operations play a relevant role in Hebbian training.

\section{Conclusions and future work}
\label{sec:conclusions}

In summary, our results suggest that our semi-supervised approach leveraging Hebbian learning is preferable w.r.t. backprop training (from scratch or with VAE pre-training) to perform training in low sample efficiency regimes where only a limited number of labeled samples is available. Specifically, our results on CIFAR10, CIFAR100, and Tiny ImageNet show that HPCA performs better than backprop training (from scratch or with VAE pre-training) in sample efficiency regimes in which only a small portion of the training set (between 1\% and 5\%) is assumed to be labeled. In addition, The HPCA+FT approach helps to further improve performance. Therefore, our method is preferable in scenarios in which manually labeling a large number of training samples would be too expensive, while gathering unlabeled samples is relatively cheap.

In future work, further improvements might come from exploring more complex feature extraction strategy, which can also be formulated as Hebbian learning variants, such as Kernel-PCA \citep{scholkopf1998} and Independent Component Analysis (ICA) \citep{hyvarinen}. 
In addition, it would be interesting to replicate this work also to the context of Spiking Neural Networks (SNNs), where the Hebbian principle is implemented by the Spike Timing Dependent Plasticity (STDP) learning rule \citep{gerstner}. SNNs are more realistic models of biological neural computation, which use pulses (called \textit{spikes}) to encode signals, rather than continuous values. This communication paradigm is the key towards energy-efficient computation in the brain \citep{javed2010}, and is being currently implemented in \textit{neuromorphic hardware} \citep{furber2014, wu2015}. In this scenario, it is necessary to map the variants of the Hebbian rule to corresponding STDP variants and test their effectiveness for SNN training. 
Finally, an exploration on the behavior of such algorithms w.r.t. adversarial examples also deserves attention.

\section{Acknowledgements}
This work was partially supported by the H2020 project AI4EU under GA 825619 and by the H2020 project AI4Media under GA 951911.


\bibliographystyle{elsarticle-num} 
\bibliography{references.bib}

\begin{thebibliography}{10}
\expandafter\ifx\csname url\endcsname\relax
  \def\url#1{\texttt{#1}}\fi
\expandafter\ifx\csname urlprefix\endcsname\relax\def\urlprefix{URL }\fi
\expandafter\ifx\csname href\endcsname\relax
  \def\href#1#2{#2} \def\path#1{#1}\fi

\bibitem{he2016}
K.~He, X.~Zhang, S.~Ren, J.~Sun, Deep residual learning for image recognition,
  in: Proceedings of the IEEE conference on computer vision and pattern
  recognition, 2016, pp. 770--778.

\bibitem{silver2016}
D.~Silver, A.~Huang, C.~J. Maddison, A.~Guez, L.~Sifre, G.~Van Den~Driessche,
  J.~Schrittwieser, I.~Antonoglou, V.~Panneershelvam, M.~Lanctot, et~al.,
  Mastering the game of go with deep neural networks and tree search, nature
  529~(7587) (2016) 484.

\bibitem{devlin2018}
J.~Devlin, M.-W. Chang, K.~Lee, K.~Toutanova, Bert: Pre-training of deep
  bidirectional transformers for language understanding, arXiv preprint
  arXiv:1810.04805 (2018).

\bibitem{bengio2007}
Y.~Bengio, P.~Lamblin, D.~Popovici, H.~Larochelle, Greedy layer-wise training
  of deep networks, in: Advances in neural information processing systems,
  2007, pp. 153--160.

\bibitem{larochelle2009}
H.~Larochelle, Y.~Bengio, J.~Louradour, P.~Lamblin, Exploring strategies for
  training deep neural networks., Journal of machine learning research 10~(1)
  (2009).

\bibitem{weston2012}
J.~Weston, F.~Ratle, H.~Mobahi, R.~Collobert, Deep learning via semi-supervised
  embedding, in: Neural networks: Tricks of the trade, Springer, 2012, pp.
  639--655.

\bibitem{kingma2014}
D.~P. Kingma, S.~Mohamed, D.~Jimenez~Rezende, M.~Welling, Semi-supervised
  learning with deep generative models, Advances in neural information
  processing systems 27 (2014) 3581--3589.

\bibitem{rasmus2015}
A.~Rasmus, M.~Berglund, M.~Honkala, H.~Valpola, T.~Raiko, Semi-supervised
  learning with ladder networks, in: Advances in neural information processing
  systems, 2015, pp. 3546--3554.

\bibitem{zhang2016}
Y.~Zhang, K.~Lee, H.~Lee, Augmenting supervised neural networks with
  unsupervised objectives for large-scale image classification, in:
  International conference on machine learning, 2016, pp. 612--621.

\bibitem{chen2020}
T.~Chen, S.~Kornblith, M.~Norouzi, G.~Hinton, A simple framework for
  contrastive learning of visual representations, in: International conference
  on machine learning, PMLR, 2020, pp. 1597--1607.

\bibitem{haykin}
S.~Haykin, Neural networks and learning machines, 3rd Edition, Pearson, 2009.

\bibitem{gerstner}
W.~Gerstner, W.~M. Kistler, Spiking neuron models: Single neurons, populations,
  plasticity, Cambridge university press, 2002.

\bibitem{oreilly}
R.~C. O'Reilly, Y.~Munakata, Computational explorations in cognitive
  neuroscience: Understanding the mind by simulating the brain, MIT press,
  2000.

\bibitem{kingma2013}
D.~P. Kingma, M.~Welling, Auto-encoding variational bayes, arXiv preprint
  arXiv:1312.6114 (2013).

\bibitem{weston2014}
J.~Weston, S.~Chopra, A.~Bordes, Memory networks, arXiv preprint
  arXiv:1410.3916 (2014).

\bibitem{grossberg1976a}
S.~Grossberg, Adaptive pattern classification and universal recoding: I.
  parallel development and coding of neural feature detectors, Biological
  cybernetics 23~(3) (1976) 121--134.

\bibitem{kohonen1982}
T.~Kohonen, Self-organized formation of topologically correct feature maps,
  Biological cybernetics 43~(1) (1982) 59--69.

\bibitem{sanger1989}
T.~D. Sanger, Optimal unsupervised learning in a single-layer linear
  feedforward neural network, Neural networks 2~(6) (1989) 459--473.

\bibitem{karhunen1995}
J.~Karhunen, J.~Joutsensalo, Generalizations of principal component analysis,
  optimization problems, and neural networks, Neural Networks 8~(4) (1995)
  549--562.

\bibitem{becker1996a}
S.~Becker, M.~Plumbley, Unsupervised neural network learning procedures for
  feature extraction and classification, Applied Intelligence 6~(3) (1996)
  185--203.

\bibitem{pehlevan2015a}
C.~Pehlevan, T.~Hu, D.~B. Chklovskii, A hebbian/anti-hebbian neural network for
  linear subspace learning: A derivation from multidimensional scaling of
  streaming data, Neural computation 27~(7) (2015) 1461--1495.

\bibitem{pehlevan2015c}
C.~Pehlevan, D.~B. Chklovskii, Optimization theory of hebbian/anti-hebbian
  networks for pca and whitening, in: 2015 53rd Annual Allerton Conference on
  Communication, Control, and Computing (Allerton), IEEE, 2015, pp. 1458--1465.

\bibitem{wadhwa2016a}
A.~Wadhwa, U.~Madhow, Learning sparse, distributed representations using the
  hebbian principle, arXiv preprint arXiv:1611.04228 (2016).

\bibitem{wadhwa2016b}
A.~Wadhwa, U.~Madhow, Bottom-up deep learning using the hebbian principle
  (2016).

\bibitem{bahroun2017}
Y.~Bahroun, A.~Soltoggio, Online representation learning with single and
  multi-layer hebbian networks for image classification, in: International
  Conference on Artificial Neural Networks, Springer, 2017, pp. 354--363.

\bibitem{krotov2019}
D.~Krotov, J.~J. Hopfield, Unsupervised learning by competing hidden units,
  Proceedings of the National Academy of Sciences 116~(16) (2019) 7723--7731.

\bibitem{hwta}
G.~Amato, F.~Carrara, F.~Falchi, C.~Gennaro, G.~Lagani, Hebbian learning meets
  deep convolutional neural networks, in: International Conference on Image
  Analysis and Processing, Springer, 2019, pp. 324--334.

\bibitem{thesis}
G.~Lagani, Hebbian learning algorithms for training convolutional neural
  networks, Master's thesis, School of Engineering, University of Pisa, Italy
  (2019).

\bibitem{magotra2019}
A.~Magotra, J.~kim, Transfer learning for image classification using hebbian
  plasticity principles, in: Proceedings of the 2019 3rd International
  Conference on Computer Science and Artificial Intelligence, 2019, pp.
  233--238.

\bibitem{magotra2020}
A.~Magotra, J.~Kim, Improvement of heterogeneous transfer learning efficiency
  by using hebbian learning principle, Applied Sciences 10~(16) (2020) 5631.

\bibitem{canto2020}
F.~J.~A. Canto, Convolutional neural networks with hebbian-based rules in
  online transfer learning, in: Mexican International Conference on Artificial
  Intelligence, Springer, 2020, pp. 35--49.

\bibitem{yosinski2014}
J.~Yosinski, J.~Clune, Y.~Bengio, H.~Lipson, How transferable are features in
  deep neural networks?, arXiv preprint arXiv:1411.1792 (2014).

\bibitem{foldiak1990}
P.~F{\"o}ldiak, Forming sparse representations by local anti-hebbian learning,
  Biological cybernetics 64~(2) (1990) 165--170.

\bibitem{olshausen1996a}
B.~A. Olshausen, D.~J. Field, Emergence of simple-cell receptive field
  properties by learning a sparse code for natural images, Nature 381~(6583)
  (1996) 607.

\bibitem{cifar}
A.~Krizhevsky, G.~Hinton, Learning multiple layers of features from tiny images
  (2009).

\bibitem{tinyimagenet}
J.~Wu, Q.~Zhang, G.~Xu, Tiny imagenet challenge, Tech. rep., Technical report,
  Stanford University, 2017. Available online at http~… (2017).

\bibitem{krizhevsky2012}
A.~Krizhevsky, I.~Sutskever, G.~E. Hinton, Imagenet classification with deep
  convolutional neural networks, Advances in neural information processing
  systems (2012).

\bibitem{higgins2016}
I.~Higgins, L.~Matthey, A.~Pal, C.~Burgess, X.~Glorot, M.~Botvinick,
  S.~Mohamed, A.~Lerchner, beta-vae: Learning basic visual concepts with a
  constrained variational framework (2016).

\bibitem{agrawal2014}
P.~Agrawal, R.~Girshick, J.~Malik, Analyzing the performance of multilayer
  neural networks for object recognition, arXiv preprint arXiv:1407.1610
  (2014).

\bibitem{scholkopf1998}
B.~Sch{\"o}lkopf, A.~Smola, K.-R. M{\"u}ller, Nonlinear component analysis as a
  kernel eigenvalue problem, Neural computation 10~(5) (1998) 1299--1319.

\bibitem{hyvarinen}
A.~Hyvarinen, J.~Karhunen, E.~Oja, Independent component analysis, Studies in
  informatics and control 11~(2) (2002) 205--207.

\bibitem{javed2010}
F.~Javed, Q.~He, L.~E. Davidson, J.~C. Thornton, J.~Albu, L.~Boxt, N.~Krasnow,
  M.~Elia, P.~Kang, S.~Heshka, et~al., Brain and high metabolic rate organ
  mass: contributions to resting energy expenditure beyond fat-free mass, The
  American journal of clinical nutrition 91~(4) (2010) 907--912.

\bibitem{furber2014}
S.~B. Furber, F.~Galluppi, S.~Temple, L.~A. Plana, The spinnaker project,
  Proceedings of the IEEE 102~(5) (2014) 652--665.

\bibitem{wu2015}
X.~Wu, V.~Saxena, K.~Zhu, S.~Balagopal, A cmos spiking neuron for
  brain-inspired neural networks with resistive synapses andin situlearning,
  IEEE Transactions on Circuits and Systems II: Express Briefs 62~(11) (2015)
  1088--1092.

\end{thebibliography}

\end{document}